%% file: paper.tex
\documentclass[]{jingdong}
\usepackage[toc,page,header]{appendix}
\input{sections/common}

\usepackage{cleveref}
\theoremstyle{plain}

\theoremstyle{definition}

\theoremstyle{remark}
\usepackage{subcaption}

\usepackage[textsize=tiny]{todonotes}
\usepackage{makecell}
\usetikzlibrary{arrows.meta,positioning,shapes.geometric,calc,fit,backgrounds}

\usepackage{float}
\usepackage{tabularx}
\usepackage{bm}
\usepackage{wrapfig}
\usepackage{caption}

\title{\textbf{Oxygen-TryOn: Fashion-Native Foundation Model \\for Any-item Virtual Try-On}}
\author{Oxygen AIGC Group \& Joy Future Academy, JD}

\input{sections/abstract}

\protected\def\jdapplogo{\raisebox{-0.3\height}{\includegraphics[height=1.7em]{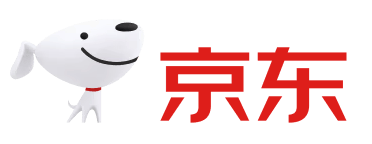}}}
\checkdata[Project Page]{\url{https://oxygenvision.github.io/Oxygen-TryOn/}}
\checkdata[\jdapplogo~Application]{\href{https://mini-app-static.jd.com/apps/mpshare/index.html?category=jump\&des=jdmp\&vapptype=1\&appId=C47CCEE416A34461C62A66A94E224594\&path=\&scene=uywxna}{JINGDONG TryOn}}
\begin{document}
\maketitle

\input{sections/intro}
\input{sections/data}

\input{sections/model}
\input{sections/training}

\input{sections/experiments}

\input{sections/application}

\input{sections/related_work}
\input{sections/limitation}
\input{sections/conclusion}
\input{sections/acknowledge}
\input{sections/showcase}
\clearpage
\bibliographystyle{plainnat}
\bibliography{cite}

% \clearpage
% \beginappendix
% \input{sections/appendix}
\end{document}

%% file: sections/common.tex
\usepackage{latexsym}
\usepackage[T1]{fontenc}
\usepackage[utf8]{inputenc}
\usepackage{microtype}
\usepackage{inconsolata}
\usepackage{graphicx}
\usepackage{hyperref}       % hyperlinks
\usepackage{url}            % simple URL typesetting
\usepackage{booktabs}       % professional-quality tables
\usepackage{amsfonts}       % blackboard math symbols
\usepackage{nicefrac}       % compact symbols for 1/2, etc.
\usepackage{stackengine}
\usepackage{microtype}      % microtypography
\usepackage{colortbl}
\usepackage{xcolor}
\usepackage{amsmath}
\usepackage{amssymb}
\usepackage{amsthm}
\usepackage{mathrsfs}
\usepackage{pifont}
\usepackage{MnSymbol}
\usepackage{balance}
\usepackage{enumitem}
\usepackage{listings}
\usepackage{xcolor}
\usepackage{natbib}
\usepackage{multicol}
\AtBeginDocument{%
  \providecommand\BibTeX{{%
    \normalfont B\kern-0.5em{\scshape i\kern-0.25em b}\kern-0.8em\TeX}}}

\makeatletter
\DeclareRobustCommand\onedot{\futurelet\@let@token\@onedot}
\def\@onedot{\ifx\@let@token.\else.\null\fi}
\newcommand{\eg}{\emph{e.g\@\onedot}}

\usepackage{setspace}
\usepackage{mathtools}

\usepackage{multirow,booktabs}
\usepackage{subcaption}

\newcommand{\owo}[1]{\textsc{OAgents}}

\definecolor{lightgreen}{RGB}{144, 238, 144} 
\definecolor{lightred}{RGB}{255, 105, 97}

\newtcolorbox{promptbox}[2][Prompt]{
colback=black!5!white,
arc=5pt, 
boxrule=0.5pt,
fonttitle=\bfseries,
title=#1, 
before upper={\small}, fontupper=\fontfamily{ptm}\selectfont,
colframe=#2, % 使用传递的参数来设定 colframe
}
\definecolor{ogreen}{RGB}{34, 139, 34}

\usepackage{color}
\usepackage{multirow}
\usepackage{booktabs}
\usepackage{wrapfig}
\usepackage{soul}
\usepackage{colortbl}
\usepackage{bbding}
\usepackage[ruled,linesnumbered]{algorithm2e}
\definecolor{mygray}{gray}{.9}
\definecolor{mypink}{rgb}{.99,.5,.5}
\definecolor{mycyan}{cmyk}{.3,0,0,0}

\definecolor{headerpink}{HTML}{FDECEC}   % 类别标题1-淡粉色
\definecolor{headerblue}{HTML}{E8F2FF}   % 类别标题2-淡蓝色
\definecolor{rowyellow}{HTML}{FFFBE6}    % "Ours" 重点行-淡黄色
\definecolor{posgreen}{HTML}{1A801A}     % 正值的绿色
\definecolor{negred}{HTML}{D10000}       % 负值的红色

%% file: sections/abstract.tex
\abstract{

We present \textbf{Oxygen-TryOn}, a unified foundation model for \emph{any-item} virtual try-on. Rather than repurposing a general-purpose image editor
  through prompting---an approach that tends to hallucinate garment details, drift on subject identity, and break down on fine-grained texture---Oxygen-TryOn is \emph{fashion-native}: it is purpose-built for try-on, powered by a dedicated data engine and try-on-specific training.
  Given one or more reference items, provided either as clean product shots or as in-the-wild photos of someone already wearing them, together with a
  single target subject image, the model synthesizes a photorealistic image of that subject wearing the referenced items, spanning virtually any fashion
  category: clothing, outerwear, accessories, footwear, bags, and beyond.
  Most prior systems handle only a single garment category in a controlled studio setting, and even recent multi-reference methods remain
  garment-centric. In contrast, Oxygen-TryOn supports diverse items and real-world scenarios, including full- and half-body views, a variable number of
  references, and free multi-item composition, while faithfully preserving both the subject's identity and each reference item's fine-grained appearance,
  from textures and prints to logos and structural silhouettes. Instead of casting try-on as mask-based inpainting, we reformulate it as a
  multi-reference, understanding-driven generation task: the model reasons about what each item is and how it should be worn, handling occlusion,
  deformation, and layering rather than merely filling a predefined region, which in turn yields stronger generalization to unseen items and
  compositions.
  To unlock this capability, we build a dedicated data engine that collects, manufactures, annotates, and filters high-quality try-on data at scale, and
  we design a three-stage training recipe comprising continued pre-training (CPT), large-scale supervised fine-tuning (SFT), and reinforcement learning
  (RL). The RL stage is driven by a hybrid reward mechanism that integrates an in-house try-on reward model with a proprietary, rubric-guided
  general-purpose model, jointly supervising fine-grained consistency and instruction-level quality. As a by-product, the same model still follows
  general editing instructions (\eg, pose changes) within a single pass, with no task switching.
  Across public benchmarks and our in-house Oxygen-TryOn Bench, Oxygen-TryOn achieves state-of-the-art consistency and realism on single-item try-on and
  leads on multi-item try-on, matching or surpassing both leading proprietary systems (\eg, Nano Banana Pro, GPT-Image-2, Seedream5 Lite) and the
  strongest open-source models (\eg, FLUX.2). To the best of our knowledge, it is the first system to deliver any-item, multi-reference try-on at this
  level of fidelity. We further detail the recipe behind the model---the data engine and the CPT--SFT--RL pipeline---to make our design choices
  transparent and reusable.

}

%% file: sections/intro.tex
\begin{figure}[!ht]
\centering
\includegraphics[width=0.95\linewidth]{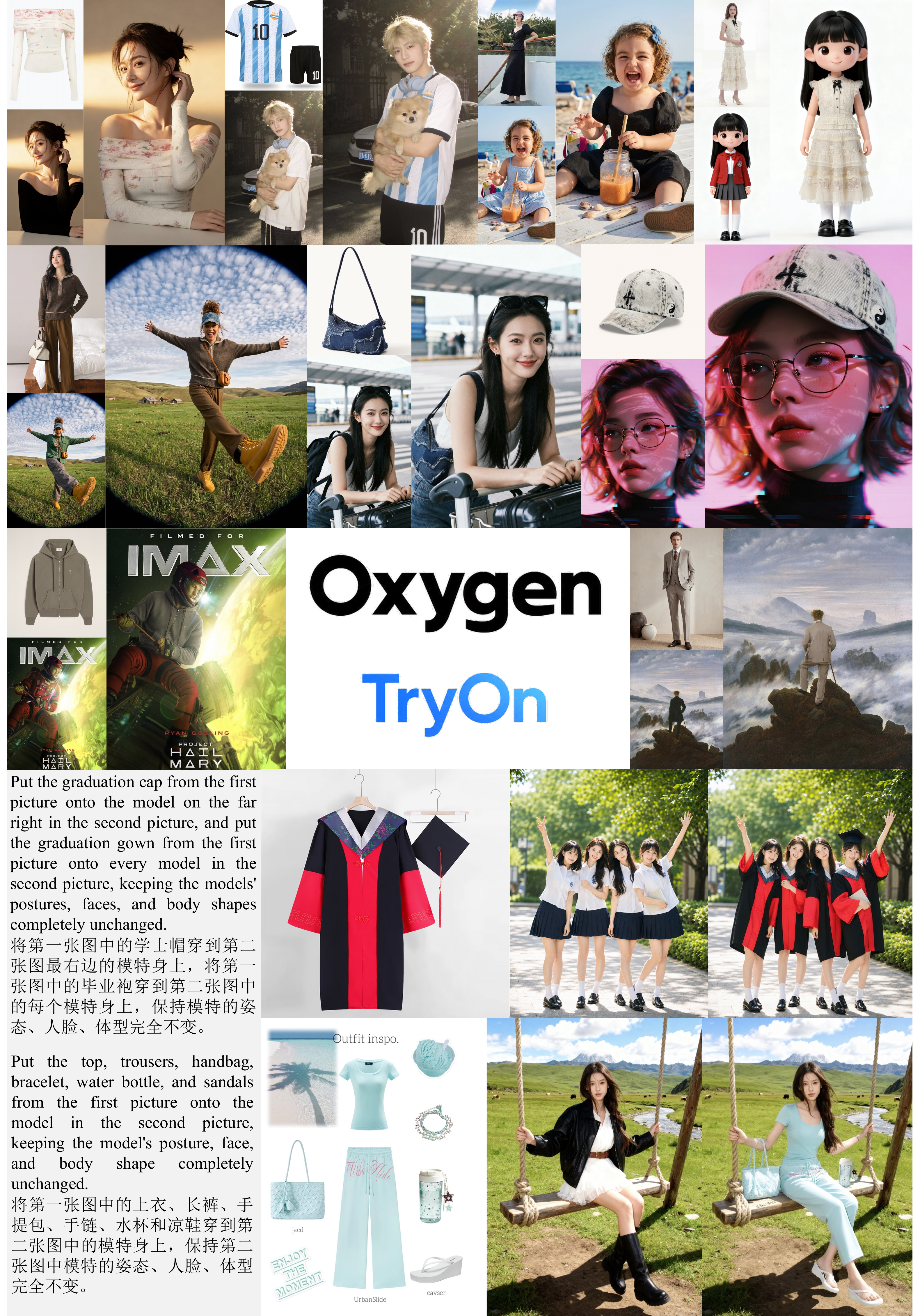}
\caption{\textbf{Oxygen-TryOn delivers high-fidelity any-item visual try-on across diverse scenarios.}}
% \caption{\textbf{Oxygen-TryOn delivers high-fidelity any-item try-on from a single reference image.} Given one reference at a time---ranging from clean product shots to in-the-wild photos of someone already wearing the item---Oxygen-TryOn dresses the target subject while faithfully preserving fine-grained item appearance (texture, structure, and logos) together with the subject's identity, body shape, and background. Beyond clothing, it transfers non-garment fashion items such as shoes, bags, and accessories, and remains robust across diverse poses and both full- and half-body views.}
% \label{fig:teaser_single}
% Given one reference image at a time, the model faithfully transfers clean garment product shots, garments shown on another person, and non-clothing fashion items such as shoes and bags onto the target subject. The cases further demonstrate strong in-the-wild generalization beyond standard fashion photography, including cross-domain targets such as movie posters, anime characters, and oil-painting-style portraits.}
\label{fig:teaser1}
\end{figure}

\begin{figure}[!ht]
\centering
\includegraphics[width=0.95\linewidth]{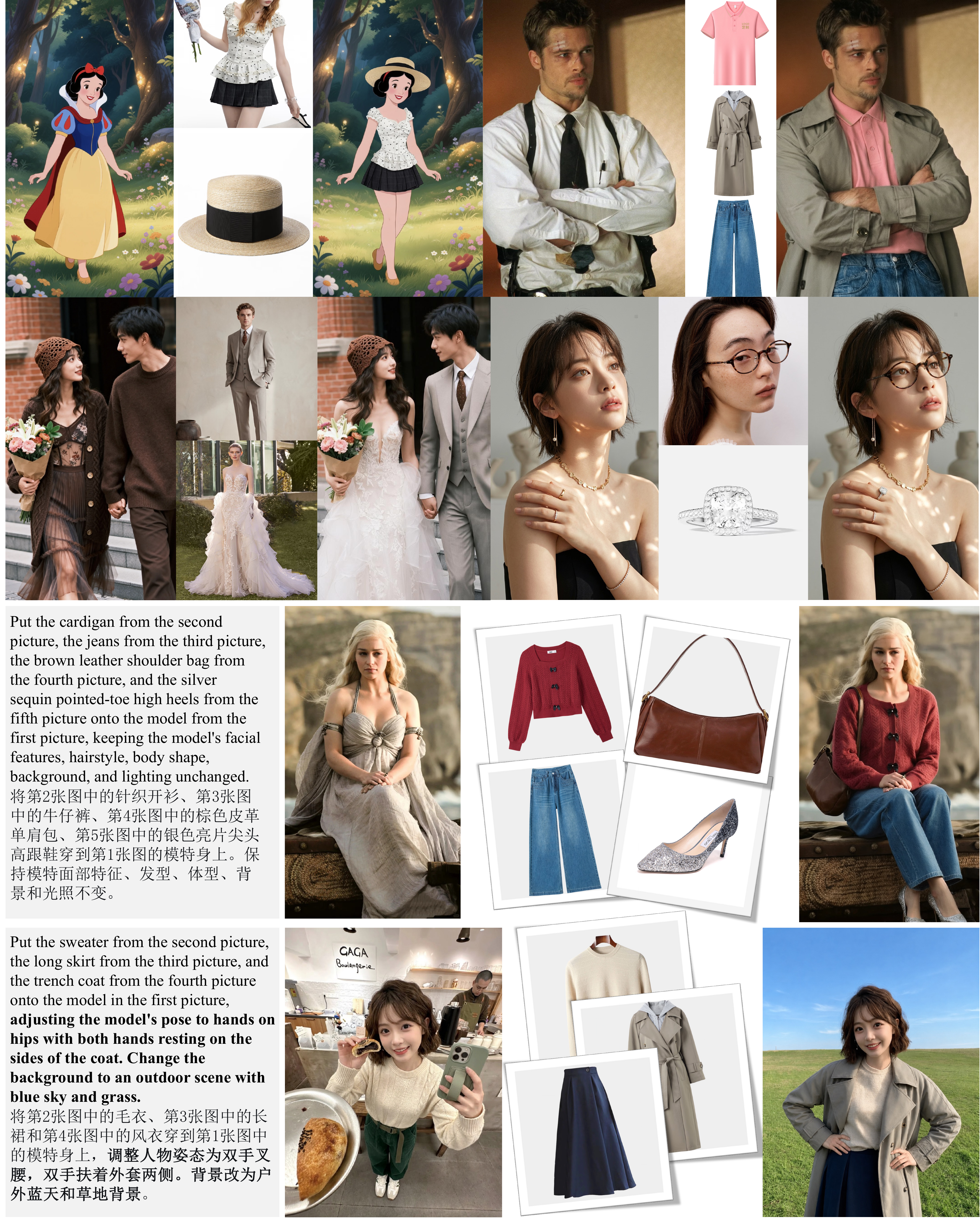}
% \caption{\textbf{Oxygen-TryOn handles challenging and complex try-on scenarios}, including full-body and half-body compositions, multi-item layering, diverse poses, and non-standard subjects. }
\caption{\textbf{Oxygen-TryOn composes multiple reference items across heterogeneous subjects, styles, and domains.} From several reference items provided at once, the model performs free multi-item composition with plausible layering and consistent per-item appearance. The cases span single- and multi-person targets and stylized cross-domain subjects such as anime characters and movie scenes, and further show that general editing directives (\eg, pose or scene changes) can be executed within the same generation pass.}
% \label{fig:teaser_multi}
\label{fig:teaser2}
\end{figure}

\begin{figure}[t]
    \centering
    \includegraphics[width=0.95\linewidth]{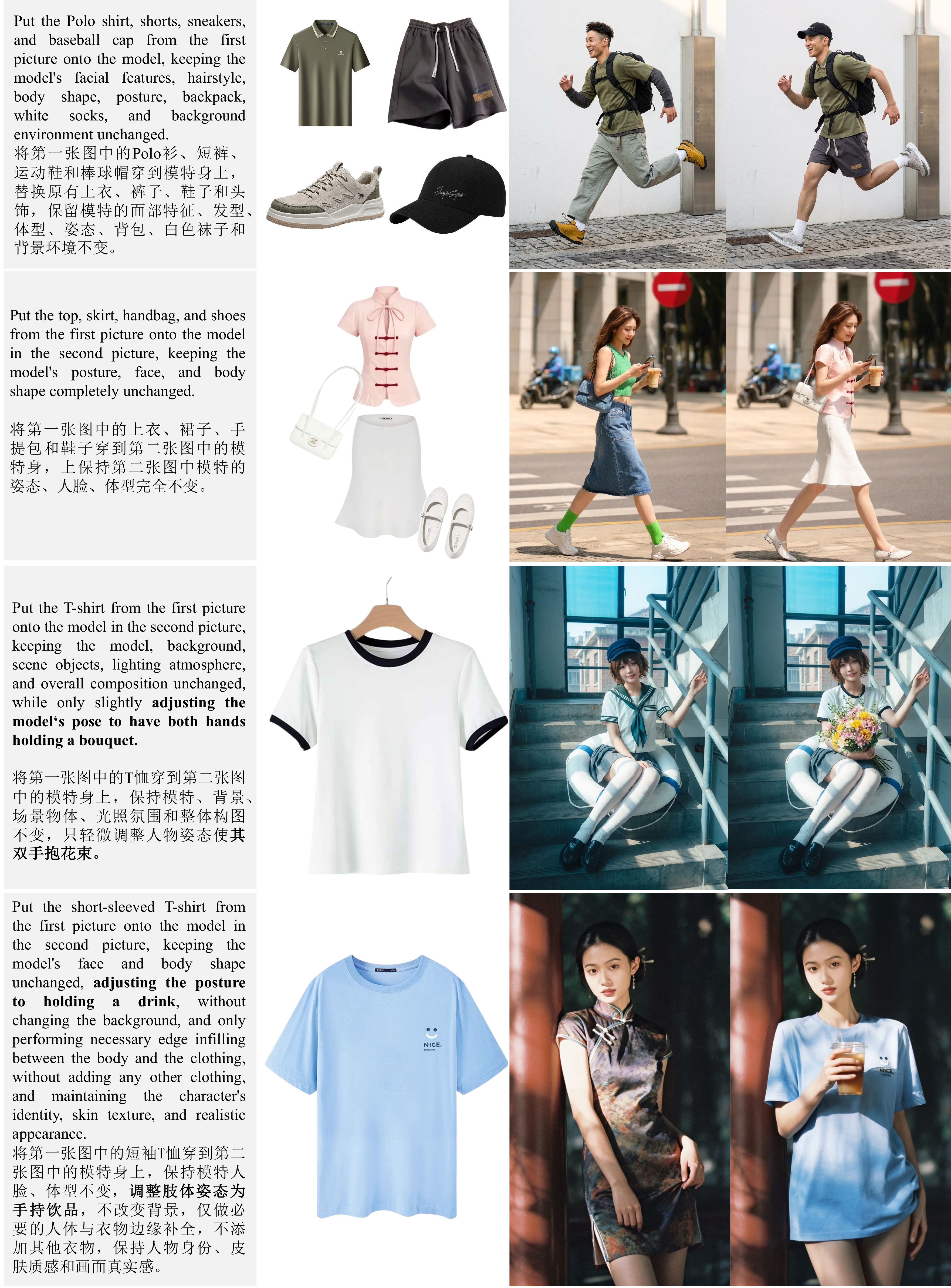}
\caption{\textbf{Oxygen-TryOn handles complex scenarios such as cluttered multi-item references and instruction-driven editing.}}
% \label{fig:teaser_complex}
\label{fig:teaser3}
\end{figure}

\clearpage

\section{Introduction}

Virtual try-on has rapidly become one of the most compelling applications of generative AI, reshaping the form of online retail where consumers can seamlessly visualize garments and accessories put on themselves when browsing~\cite{wang2018toward,han2018viton,chen2024wear,xu2024tunnel}. An ideal try-on system should accept arbitrary reference items---whether presented as clean product shots or as in-the-wild photos of someone already wearing them---faithfully preserve fine-grained item appearance and subject identity, and flexibly compose multiple items, including garments, accessories, shoes, and bags, on a single target subject under both full-body and half-body views.

Recent advances in diffusion-based generation and editing~\cite{ho2020denoising,rombach2022high,esser2024scaling} have accelerated progress toward this goal along two fronts. On one hand, powerful general-purpose generation and editing models---both proprietary~(\eg, Nano Banana Pro~\cite{google2025nanobanana}, GPT-Image-2~\cite{gptimage2_model_card}) and open-source~(\eg, Qwen-Image~\cite{wu2025qwen,qwenimage}, FLUX~\cite{flux-2-2025,blackforest2025flux2klein})---exhibit remarkable semantic understanding and high-fidelity synthesis, and serve as strong starting points for task-specific fine-tuning. On the other hand, the academic community continues to advance dedicated try-on models along complementary angles such as spatial alignment, garment warping, and identity preservation~\cite{chong2024catvtonconcatenationneedvirtual,jiang2024fitdit,xu2025ootdiffusion,zhou2024learning,guo2025any2anytryon}.

Despite this progress, building a professional \emph{any-item} try-on model remains challenging. First, most existing methods are confined to a single garment category, such as upper-body tops, and to a narrow studio setting, lacking the flexibility to support accessories, shoes, bags, and free multi-item composition~\cite{chong2024catvtonconcatenationneedvirtual,jiang2024fitdit,xu2025ootdiffusion}. Second, many methods implicitly assume that the reference is a pristine flat-lay garment; in practice, user-provided references are highly unconstrained and frequently contain complex backgrounds or in-the-wild portraits of other people, which existing models handle poorly. Third, strong \emph{consistency}---faithful preservation of garment texture, structure, and logos on one side, and of the target subject's identity, body shape, and background on the other---remains difficult to achieve, especially under diverse poses and across half-body and full-body views. As a result, even the strongest general-purpose editing models still leave a clear gap toward a unified, high-fidelity try-on solution. Crucially, the systems that come closest to closing this gap are proprietary, and the open-source community still lacks a model that matches them while exposing the full pipeline needed to reproduce and extend such capability.

In this work, we present \textbf{Oxygen-TryOn}, a unified foundation model for any-item virtual try-on. Rather than treating try-on as a constrained inpainting problem, we reformulate it as a multi-reference, understanding-driven image generation task. Formally, given a single target subject image $I_s$, a set of reference item images $\mathcal{R}=\{I_1, I_2, \dots, I_n\}$ (each depicting a garment, accessory, shoe, or bag, presented either as a clean product shot or as an in-the-wild photo of someone wearing the item), and an optional textual instruction $T$, the model synthesizes an output image $I_o$ in which the target subject wears the referenced items:
\begin{equation}
I_o = \mathcal{G}\big(I_s, \mathcal{R}, T\big).
\end{equation}
The number of reference items $n$ is variable, enabling single-item try-on as well as free multi-item composition, and the instruction $T$ may optionally carry general editing directives (f\eg, adjusting the subject's pose) that are executed within the same generation pass. Unlike mask-based inpainting formulations, this end-to-end formulation requires no explicit garment-agnostic mask, allowing the model to reason holistically about layering, occlusion, and the interaction between items and the subject.

Oxygen-TryOn is built on top of the JoyAI-Image-Edit~\cite{song2026joyai}, inheriting a strong multimodal foundation that couples a  MLLM for reference and instruction understanding with a large-scale MMDiT for high-fidelity synthesis. On this foundation, we develop a three-stage training recipe consisting of continued pre-training (CPT), large-scale supervised fine-tuning (SFT), and reinforcement learning (RL), and a dedicated data engine that collects, manufactures, annotates, and filters high-quality try-on data at scale.

We conduct extensive quantitative and qualitative evaluations across single- and multi-item scenarios, covering diverse garments, accessories, shoes, and bags under both full-body and half-body views. Across public benchmarks and our in-house Oxygen-TryOn Bench, Oxygen-TryOn achieves state-of-the-art consistency and realism, surpassing both strong proprietary systems such as Nano Banana Pro~\cite{google2025nanobanana}, GPT-Image-2~\cite{gptimage2_model_card}, and Seedream5 Lite~\cite{bytedance2026seedream}, and leading open-source models such as FLUX.2~\cite{blackforest2025flux2klein}. This advantage holds across both single-item try-on and multi-item composition. Figures~\ref{fig:teaser1} and~\ref{fig:teaser2} further illustrate its robustness across challenging in-the-wild conditions and complex compositions.

\noindent The key contributions of Oxygen-TryOn can be summarized as follows:
\begin{itemize}
    \item \textbf{A Powerful Any-Item Try-On System.} We present Oxygen-TryOn, a unified model that dresses a target subject with an arbitrary combination of garments, accessories, shoes, and bags from heterogeneous reference images (clean product shots or worn-on photos), under both full-body and half-body views. To the best of our knowledge, it is the first system to deliver any-item, multi-reference try-on at this level of fidelity. 
    % We release the model weights together with the full inference system---including the prompt-enhancer protocol that drives it---so that practitioners can directly deploy it.

    \item \textbf{A Strong Foundation with a Practical Training Recipe.} We build Oxygen-TryOn on the JoyAI-Image-Edit architecture and pretrained weights, and design a three-stage CPT--SFT--RL recipe together with a dedicated data engine for collecting, manufacturing, annotating, and filtering high-quality try-on data. We document this recipe in detail to provide a transparent, reproducible route to high-fidelity try-on models.

    \item \textbf{State-of-the-Art Consistency and Realism.} Across diverse try-on scenarios, Oxygen-TryOn attains leading item--subject consistency and photorealism, outperforming strong proprietary and open-source baselines across single-item and multi-item try-on. As a by-product, it also retains general instruction-based editing (\eg, pose changes) within the same model.
\end{itemize}

%% file: sections/data.tex
\section{Data Engine for Any-item Virtual Try-On}
\label{sec:data}
High-quality and diverse triplets (\textit{i.e.}, item images, subject images, and try-on results) are essential for training a general-purpose virtual try-on model. However, well-aligned triplets are extremely scarce in real-world scenarios, as these images are usually collected as independent visual assets. Consequently, directly obtaining large-scale supervised try-on data that covers diverse categories and scenarios is highly challenging.

\begin{figure}[t]
\centering
\includegraphics[width=\linewidth]{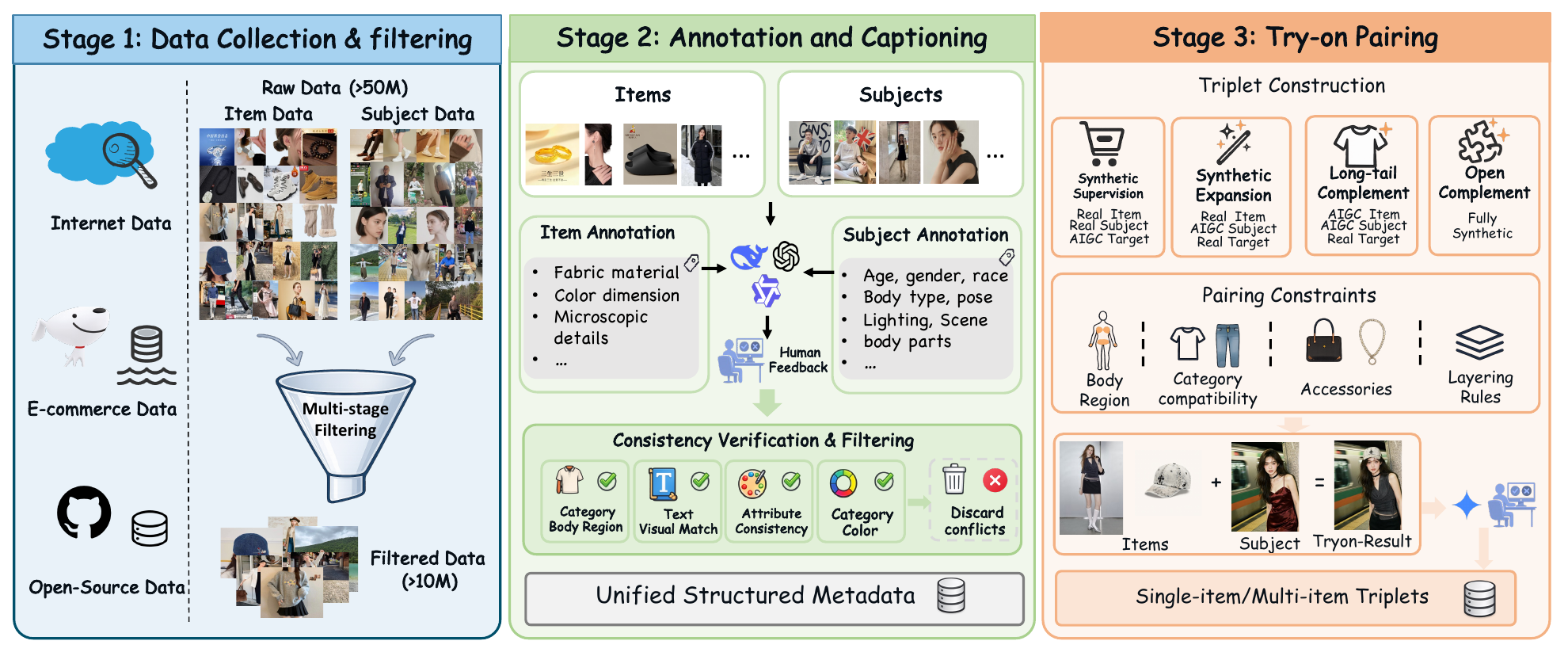}
    
\caption{\textbf{Overview of the training data engine.} The engine runs in three stages: it first \emph{collects} raw item and subject images from e-commerce and open-domain sources and applies multi-stage \emph{filtering}; it then \emph{annotates and captions} the retained images with item attributes and subject metadata; and it finally constructs item--subject--result try-on triplets through constrained \emph{pairing} before adding them to the training pool. The pool is refined iteratively with human feedback as the model improves.}
\label{fig:data_engine}
\end{figure}

To address this challenge, we build a scalable data engine that transforms heterogeneous images of fashion scenarios into a unified try-on triplet format. Each triplet consists of one or multiple reference images of the wearable item, a target subject specified by either a full-body person image or a local body region, and the resulting image depicting the subject wearing or carrying the specified item. The entire data engine consists of three main stages: data collection and filtering, annotation and captioning, and try-on pairing. 
It integrates real e-commerce images, open-domain fashion images, and model-generated samples, covering a wide range of items—garments, shoes, bags, hats, glasses, jewelry, and other accessories—as well as diverse body shapes, poses, identities, styles, and imaging conditions. The overall pipeline is illustrated in Figure~\ref{fig:data_engine}.

\subsection{Data Collection \& Filtering}

\textbf{Multi-source Data Collection:} Unlike conventional virtual try-on datasets, which primarily focus on specific scenarios such as upper-body garments, lower-body garments, or full-body outfits, our dataset covers a much broader spectrum of wearable products. The reference items include not only standard clothing, but also shoes, bags, hats, glasses, jewelry, and other accessories. Correspondingly, the target subjects are not limited to full-body person images; they also encompass half-body portraits and localized body-part images, such as feet, hands, heads, upper bodies, and lower bodies. This design enables the dataset to support both conventional garment try-on and more fine-grained wearable-product transfer tasks, including shoe try-on, ring try-on, hat and glasses try-on, and bag-and-accessory composition.

We construct the raw data pool from three complementary sources. The first source is real e-commerce data, including clean product images, white-background product images, product-detail images, and model-worn images. These data provide high-fidelity product appearance and realistic wearing examples. The second source is open-domain fashion imagery, which improves the diversity of human identities, poses, backgrounds, styles, and imaging conditions. The third source is model-generated synthetic data, which is used to expand long-tail categories, local body regions, rare wearing combinations, and challenging pose-view configurations that are underrepresented in real data. By integrating these three sources, we build a diverse candidate data pool containing product images, target-subject images, and dressed-person images.

% \subsection{Data Manufacturing}
% \label{sec:data_manufacture}

\textbf{Data Filtering:}
To ensure data quality, we perform fine-grained cleaning and filtering on over 50 million raw images. Specifically, we progressively filter the data from multiple perspectives, including single-image quality, image content, item-subject matching, and final triplet consistency, thereby removing low-quality, weakly related, or unreliable training samples.

\noindent-- \textit{Basic Filtering:} We first remove corrupted files, images below the minimum resolution threshold of 1024p, duplicated samples identified by MD5 and semantic similarity, and unsafe images with explicit content. This step eliminates invalid or low-quality samples from the raw data pool.

\noindent-- \textit{Rule-based Hard Filtering:} We further discard samples according to a set of low-level image statistics and heuristic rules, including image resolution, aspect ratio, abnormal white margins, solid-color borders, contrast, sharpness, brightness, entropy, and saturation.

\noindent-- \textit{Text and Overlay Filtering:} We use VLM\cite{qwen3vl2025,qwen2.5vl} and OCR models\cite{cui2025paddleocr30technicalreport,cui2026paddleocr}, together with a patch-detection model, to remove images with obvious graphic patches, watermarks, or severe textual interference.

Beyond the above standard filters, we design two task-specific filters that are particularly effective for try-on.

\noindent-- \textit{Person and Item Validity Filtering:} We use recognition and detection models to verify whether an image contains the expected visual content. For dressed-person images and target-subject images, we require the presence of a valid human body or a valid target body region. For clean product images, we remove samples that unexpectedly contain human figures, so that pure product images can be clearly separated from scene images or worn-product images. This prevents ambiguous supervision signals during training.

\noindent-- \textit{Aesthetic and Perceptual Quality Filtering:} To further improve subjective visual quality and overall usability, we introduce multiple complementary image-quality and aesthetic scoring models, including Aesthetic-Predictor-v2.5\cite{discus2024aestheticpredictorv25}, KONIQ\cite{hosu2020koniq}, SPAQ\cite{fang2020perceptual} and APDD\cite{jin2024apddv2} related dimensions. These models assess images from different perspectives, such as aesthetic preference, perceptual quality, composition layout, lighting and color, detail completeness, and overall visual appeal. Specifically, we discretize the continuous score of each model into five levels: \textit{bad}, \textit{poor}, \textit{fair}, \textit{good}, and \textit{excellent}. We then combine majority voting with weighted scoring to obtain the final quality score. This strategy reduces the bias of individual scoring models and yields more stable filtering decisions, allowing us to remove samples with poor aesthetics, cluttered composition, or unsatisfactory visual quality.

% Because naturally paired try-on data is scarce, the core of our data engine is a \emph{data manufacturing} pipeline that synthesizes high-quality (reference item, dressed-person) pairs at scale. \TODO{Describe the full data-manufacturing pipeline. This is expected to cover: (1) how reference items are decomposed/extracted from dressed-person images (segmentation, matting, retrieval-based recall of the matching product); (2) how clean product shots and worn-on photos are paired with target persons; (3) how multi-item layered outfits are composed under physically plausible coexistence rules; (4) any model-in-the-loop generation/augmentation used to create training targets and the self-improving loop that bootstraps higher-quality data from earlier model checkpoints; and (5) measures taken to avoid leakage and trivial copy shortcuts.}

% \noindent\textbf{Placeholder summary.} At a high level, the manufacturing pipeline produces three families of samples: (a) \emph{product-to-person} pairs, where one or more clean product shots are associated with a person wearing the corresponding items; (b) \emph{person-to-person} pairs, where items worn by one individual are transferred to a different target model; and (c) \emph{multi-item composition} pairs, where several references are combined into a single coherent outfit following layering and occlusion constraints. The exact algorithms, tools, and ratios for each family will be detailed in the final version.

\subsection{Annotation and Captioning}

After data filtering, we perform structured annotation and caption generation for the retained images, converting raw images into unified intermediate representations for pairing, training, and statistical analysis. We combine VLM, detection models, and rule-based verification to generate multi-level image descriptions for product images and target-subject images, producing diverse textual representations at different granularities. 

% We adopt Qwen3-VL-8B-Instruct as the unified captioning backbone, and generate both Chinese and English descriptions for all images to support multilingual generation.

For product images, we annotate the wearable-product category, the compatible body region, and basic visual attributes, such as color, material, texture, design, and style. To enhance fine-grained controllability, we define category-specific attribute spaces for different product types and construct a hierarchical product taxonomy through manual curation.

For target-subject images, we annotate the visible body regions, human pose, viewpoint, apparent gender, apparent age range, body-shape characteristics, and background complexity. For local body-region images, such as hands, feet, heads, upper bodies, and lower bodies, we further annotate the corresponding body-region type and the range of compatible wearable products. These annotations are used to constrain the subsequent pairing process and avoid mismatches between product categories and target subject regions.

To enhance annotation reliability, we employ multiple complementary models to generate candidate labels and produce bilingual (Chinese and English) descriptions for all images to facilitate multilingual generation. Subsequently, we apply rule-based consistency filtering. Specifically, we verify whether the product category matches the body region, whether textual entities are consistent with visual recognition results, whether key product attributes conflict with each other, and whether the caption contains category or color descriptions that are inconsistent with the image. Samples with obvious conflicts are discarded. Finally, each image is assigned unified structured metadata. Overall, the multi-level descriptions provide comprehensive textual supervision for Oxygen-TryOn, enabling the model to learn both holistic scene understanding and fine-grained attribute control.

% Each manufactured sample is enriched with structured annotations and natural-language wearing instructions. Annotations include item category and fine-grained sub-attributes, the target body region for each item, and the layering relationship among items. A customized vision-language captioner generates a precise wearing instruction that describes the intended try-on operation (\eg, which item goes where, and any general editing directives such as a pose change), which serves as the textual condition $T$ during training. \TODO{Specify the captioner used, the instruction template/format, and how editing directives (e.g., pose change) are sampled and balanced against pure try-on instructions.}

\subsection{Try-on Pairing}

Since real-world try-on triplets are difficult to obtain directly, we construct item–subject–result triplets from our cleaned and annotated item and subject libraries by generating the missing components. To achieve this while preserving realistic item details, diversifying the distribution of target subjects, supplementing long-tail garment categories, and improving the model’s generalization ability in open-domain scenarios, we design four complementary construction pipelines.

% Based on the annotated item library and subject library, we construct try-on triplets using multiple complementary pairing strategies. Our overall goal is to preserve realistic item details, diversify the distribution of target subjects, supplement long-tail garment categories, and improve the model’s generalization ability in open-domain scenarios.

% To this end, we design four complementary triplet-construction pipelines: 

% Specifically, the proposed pipelines are: a real-product-driven real-supervision pipeline, a real-product-driven synthetic-expansion pipeline, a synthetic-product-driven long-tail-complement pipeline, and a fully synthetic open-composition pipeline. The real-supervision pipeline is mainly used to preserve authentic e-commerce product textures, materials, and wearing effects. The synthetic-expansion pipeline increases the coverage of real products across diverse identities, poses, body shapes, and scenes. The long-tail-complement pipeline supplements rare categories, local body regions, and uncommon wearing relationships. The fully synthetic pipeline constructs open-composition scenarios that are difficult to collect at scale from real data. Through these complementary pipelines, real samples provide reliable supervision and product-detail constraints, while synthetic samples expand the data distribution and improve long-tail coverage.

Specifically, we construct four complementary data pipelines with different roles. The first pipeline builds training pairs from real products and real dressed-person images, providing reliable supervision for preserving authentic e-commerce textures, materials, and wearing effects. The second pipeline uses real products with synthesized target appearances to expand the coverage of identities, poses, body shapes, and scenes while maintaining product fidelity. The third pipeline introduces synthesized products to supplement rare categories, local body regions, and uncommon wearing relationships that are underrepresented in real data. The fourth pipeline generates fully synthetic compositions to cover open-ended try-on scenarios that are difficult to collect at scale from real-world sources. Together, these pipelines combine the reliability of real samples with the diversity and scalability of synthetic data, enabling both fine-grained product preservation and broad distributional coverage.

To support omnipotent try-on with multiple reference images, we further construct multi-item triplets. Multiple wearable products are not combined randomly; instead, they are sampled under constraints based on body regions, category compatibility, and physical coexistence. For example, tops, bottoms, shoes, bags, and accessories can jointly form a complete outfit, while mutually exclusive items occupying the same body region are generally not allowed to co-occur unless they form a plausible layered structure. For accessories such as hats, glasses, jewelry, and bags, we also impose constraints according to their corresponding body regions and wearing manners. This multi-item pairing strategy enables the model to learn spatial relationships, layering relationships, and overall outfit compatibility among multiple reference items, thereby supporting unified modeling from single-item try-on to full-outfit generation.

% In addition, we introduce strict constraints during pairing and data splitting to reduce data leakage and shortcut learning. We perform near-duplicate removal at the product, subject, and result levels to reduce the risk of memorizing training samples. We also remove samples where the reference item image and the result image are overly similar in pixel layout, preventing the model from exploiting trivial copy shortcuts instead of learning the intended try-on transformation. Together, these strategies improve the effectiveness of the dataset and make it more suitable for training and evaluating general-purpose virtual try-on models in open-world scenarios.

% To guarantee data quality, manufactured samples pass through a multi-stage filtering pipeline that combines automated and human checks. We apply perceptual and structural metrics to reject samples with artifacts, misalignment, or identity drift, and use a knowledge-enhanced VLM to verify that (i) every reference item appears faithfully in the target image, (ii) the target person's identity, body shape, and background are preserved, and (iii) the layering and physical structure are plausible. A final round of human verification is applied to a sampled subset to calibrate the automated filters. \TODO{Report acceptance/rejection rates, the specific metrics and thresholds, and the VLM checking prompts.}

\begin{figure}[t]
\centering
\includegraphics[width=\linewidth]{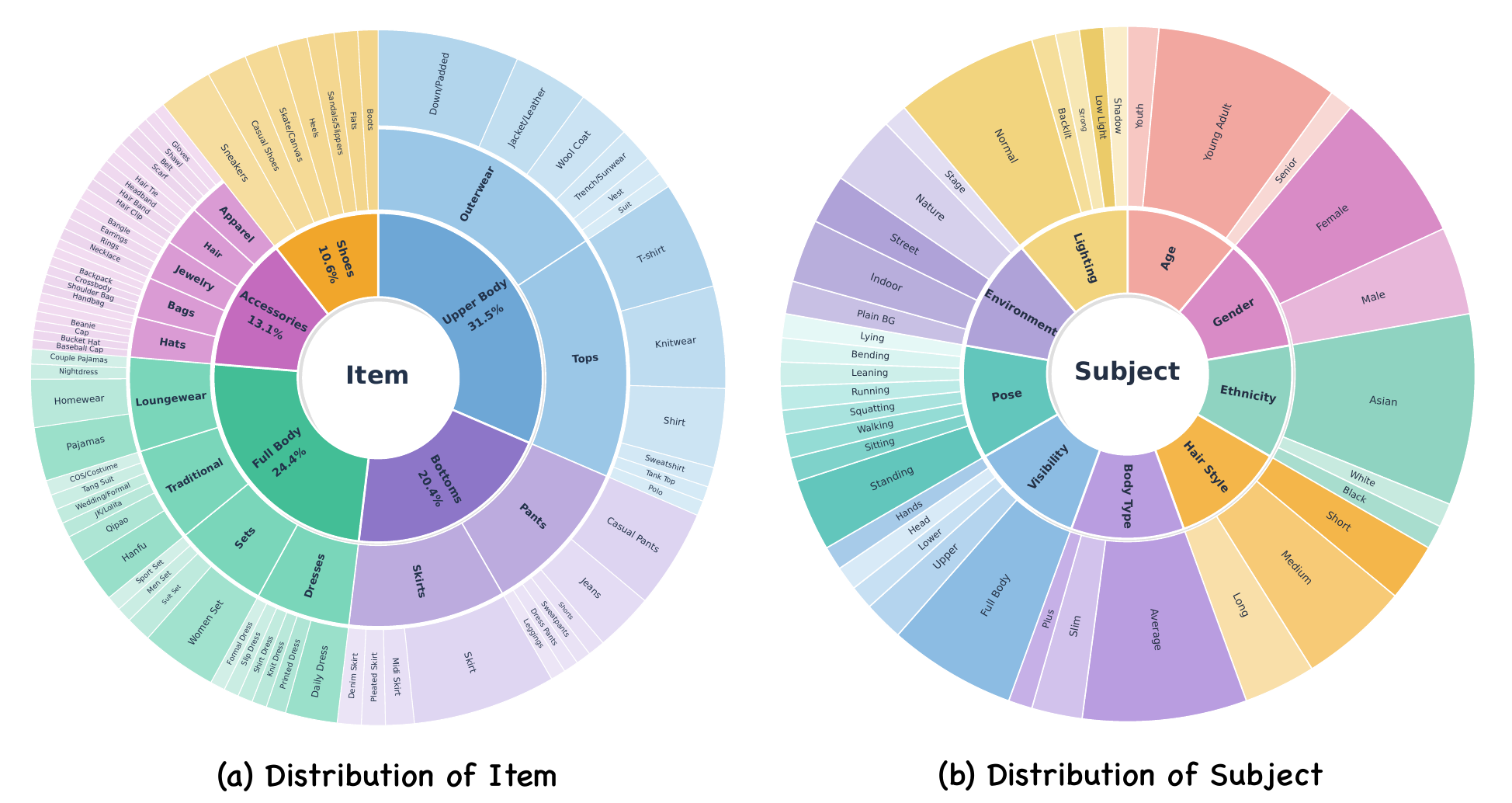}
    
\caption{\textbf{Statistics of training dataset.}
We characterize the training data from two perspectives: item categories and subject attributes. 
The item distribution covers diverse wearable items, including upper-body clothing, bottoms, full-body outfits, accessories, and shoes, with fine-grained category annotations. 
The subject distribution includes rich human-centric attributes such as age, gender, ethnicity, hairstyle, body type, visibility, pose, environment, and lighting. 
Such broad coverage enables robust and generalizable virtual try-on training.}

\label{fig:garment_model_distribution}
\end{figure}

\subsection{Dataset Overview and Statistics}

The proposed dataset is constructed for open-world virtual try-on. It is no longer limited to conventional clothing transfer, such as tops, bottoms, or full-body outfits, but instead covers a broader range of wearable items, more diverse target subjects, and more complex outfit scenarios. We characterize the data distribution from two complementary perspectives: item categories and subject attributes, as illustrated in Figure~\ref{fig:garment_model_distribution}.

\noindent\textbf{Item Distribution:} We construct a hierarchical taxonomy of wearable items. Unlike existing try-on datasets that mainly focus on garments, our dataset covers tops, bottoms, outerwear, full-body clothing, shoes, bags, hats, glasses, jewelry, scarves, and other accessories. The taxonomy contains 207 valid category mappings, covering 58 coarse-grained product groups, 151 e-commerce third-level categories, and 104 fine-grained fashion-style categories. Such category coverage enables the dataset to support not only conventional garment try-on, but also shoe try-on, bag carrying, hat and glasses wearing, jewelry placement, and multi-item outfit composition.

\noindent\textbf{Subject Distribution:} On the target-subject side, we build a diverse subject library that reflects real user-uploaded scenarios. The subject images include full-body, half-body, and local body-region images, and cover diverse apparent gender, age range, racial appearance, body shape, figure, hairstyle, pose, viewpoint, and camera framing. The pose distribution includes not only standard standing poses, but also more challenging cases such as walking, sitting, crossed legs, selfies, occlusion, and non-frontal views. Such subject diversity enables the model to learn stable try-on transformations under conditions close to real user inputs, rather than being limited to standard studio-shot subject images.

\noindent\textbf{Outfit and Scenario Distribution:} The dataset further covers diverse outfit combinations and shooting environments. Each sample may contain either a single item or a combination of multiple products, including garments, shoes, bags, hats, glasses, and jewelry. Multi-item samples are constructed according to body-region constraints, category compatibility, and physical coexistence, avoiding unreasonable item co-occurrence while preserving common layered dressing patterns. In terms of scenes, the dataset covers indoor, outdoor, studio, street, and natural environments under different lighting conditions, enabling the model to preserve human identity, background content, and product details under complex backgrounds and non-standard imaging conditions.

Overall, compared with existing virtual try-on datasets, our training data has three key characteristics. First, it extends try-on objects from conventional garments to general wearable products. Second, it supports try-on and wearing tasks for full-body persons, half-body persons, and local body regions. Third, it unifies single-item samples, multi-item samples, real-supervision samples, and synthetic-augmentation samples under the item--subject--result triplet representation. These designs provide more comprehensive supervision for training a unified virtual try-on model and enable cross-category, cross-body-region, and open-domain generalization.

% \begin{figure}[H]
% \centering
% \includegraphics[width=0.92\linewidth]{figs/tryon_placeholder/demo_multi.pdf}
% \caption{\textbf{Representative manufactured try-on training samples.} Each sample pairs one or more reference items (clean product shots or worn-on photos) and a natural-language wearing instruction with a target person and the resulting dressed image. \textit{(Placeholder figure; to be replaced with final Oxygen-TryOn data samples.)}}
% \label{fig:data_cases}
% \end{figure}

%% file: sections/model.tex
\begin{figure}[t]
    \centering
    \includegraphics[width=\linewidth,height=0.58\textheight,keepaspectratio]{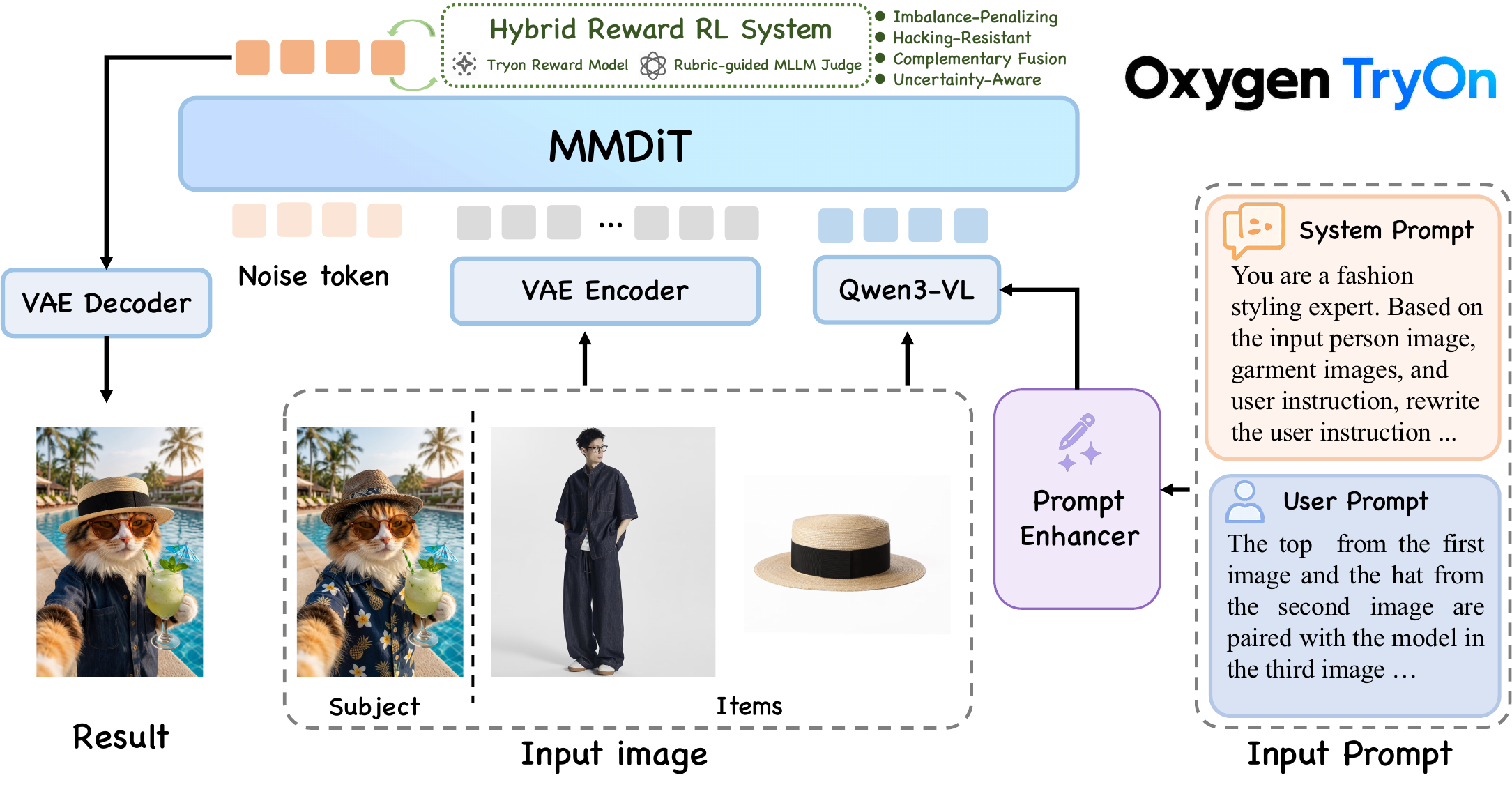}
    
    \caption{\textbf{Overall architecture of Oxygen-TryOn.} Oxygen-TryOn integrates Qwen3-VL, a VAE, and a dual-stream MMDiT for instruction-guided virtual try-on. Given a subject image, reference item images, and a user instruction, the PE module rewrites the instruction into a try-on-aware prompt. Qwen3-VL produces semantic condition tokens, while the VAE encoder extracts image latent tokens from the subject and items. The MMDiT fuses these tokens with noise tokens under timestep conditioning, and the VAE decoder generates the final dressed result. This design supports variable-length item references for both single- and multi-item try-on, and the instruction may additionally carry general editing directives (\eg, a pose change). Besides, we meticulously design a hybrid reward-driven RL system to robustly elevate the model's generative performance.}
    
\label{fig:main_arch}
\end{figure}

    \section{Model and Training}
    \label{sec:model}
    
    This section details the construction and training of Oxygen-TryOn. The model is built upon the \textbf{JoyAI-Image-Edit} foundation model~\cite{song2026joyai} and is initialized from its pretrained weights. This design enables us to inherit a strong, general-purpose multimodal prior---namely, robust scene understanding, instruction parsing, and high-fidelity synthesis---and to specialize it for any-item virtual try-on, instead of training a task-specific model from scratch. As illustrated in Figure~\ref{fig:main_arch}, the architecture combines a Multimodal Large Language Model (MLLM) with a Variational Autoencoder (VAE) and a Multimodal Diffusion Transformer (MMDiT)~\cite{esser2024scaling}. We first describe the model architecture and its adaptation to the multi-reference setting (Section~\ref{sec:model_arch}), and then present the three-stage training recipe that transforms this foundation model into an any-item try-on model (Section~\ref{sec:training}).
    
    \subsection{Model Architecture}
    \label{sec:model_arch}
    \subsubsection{Inherited Architecture}
    
    The operational workflow follows the three-stage pipeline of JoyAI-Image-Edit:
    \begin{itemize}
    \item \textbf{Multimodal Understanding:} The MLLM serves as the cognitive core. It jointly encodes the subject image, the reference item images, and the wearing instruction into a sequence of interleaved multimodal tokens, resolving which item should be placed on which body region and how items relate to one another and to the subject.
    \item \textbf{Latent Encoding:} The VAE maps the subject image and the reference images into a compact latent space suitable for diffusion modeling, while preserving fine-grained structures (\eg, fabric texture, prints, and logos) that are critical for faithful item reconstruction.
    \item \textbf{Conditional Generation:} The MMDiT serves as the generative engine. Through its dual-stream architecture, it fuses the MLLM-derived semantic priors with VAE-encoded image tokens and noise tokens, supporting high-fidelity, multi-condition synthesis of the dressed result.
    \end{itemize}
    
    \subsubsection{Multimodal Large Language Model}
    
    Following JoyAI-Image-Edit, the comprehension module is built on Qwen3-VL-8B-Instruct~\cite{qwen3vl2025} and acts as the primary interaction interface for parsing inputs and aligning modalities. Over the interleaved try-on input $(I_s, \mathcal{R}=\{I_1,\dots,I_n\}, T)$, we take the hidden states from its final layer as the primary conditioning signal for the MMDiT. This understanding-driven design is particularly important for any-item try-on, where the model must disambiguate heterogeneous references (clean product shots versus worn-on photos), associate each item with the correct body region, and respect physical layering constraints.
    
    \subsubsection{Variational Auto-Encoder and Multimodal Diffusion Transformer}
    
    A Wan-2.1-VAE~\cite{wan2025wan} compresses images into latent tokens while preserving high-frequency details that are essential for reproducing intricate garment textures and small logos. The generative core is a 16B-parameter MMDiT that jointly models the multimodal conditioning from the MLLM and the latent tokens from the VAE. To support an arbitrary number of reference items, the reference and target latents are encoded into a shared token sequence so that the MMDiT can attend across all conditions when denoising.
    
    \subsubsection{Multi-Reference Conditioning}
    \label{sec:multiref}
    
    A central design choice for any-item try-on is how to inject a variable number of heterogeneous reference items into the generative process. Oxygen-TryOn encodes the subject image and every reference image into latent tokens through the shared VAE and lays them out as a single token sequence with two functional regions: a \emph{target region}, whose tokens start as Gaussian noise and are progressively denoised into the output, and a \emph{reference region}, which carries the clean latents of all reference items and acts purely as conditioning. We deliberately avoid segment, index, or region-type embeddings, as well as any reference-tagging or image-identifier tokens; instead, the sole mechanism that organizes this sequence is the temporal ($T$) axis of the multimodal RoPE (MRoPE) position encoding. The target region and each individual reference are assigned \emph{non-overlapping intervals} along the $T$ axis, so that the MMDiT can both separate the denoising target from its conditioning and tell the references apart from their positions alone. Reference identity and placement are instead resolved on the language side: the instruction refers to each reference by its position in natural language (\eg, ``Picture~1'', ``Picture~2'') rather than through dedicated tokens, and item category is likewise conveyed only through the instruction rather than any explicit category signal. In parallel, the MLLM supplies a stream of semantic condition tokens that describe item identity and wearing intent. This design natively supports a flexible number of references and decouples item appearance (carried by VAE tokens) from item semantics and placement (carried by MLLM tokens), which we find important for preserving fine-grained details while following complex multi-item instructions.
    

%% file: sections/training.tex
\subsection{Training}
\label{sec:training}

Starting from the JoyAI-Image-Edit pretrained weights, we train Oxygen-TryOn with a three-stage pipeline: \textbf{continued pre-training (CPT)} that adapts the foundation toward the try-on domain on a balanced mixture of general-purpose and try-on data, followed by large-scale \textbf{supervised fine-tuning (SFT)} to acquire broad try-on capability, and finally \textbf{reinforcement learning (RL)} to further improve consistency, realism, and instruction following.

\subsubsection{Continued Pre-Training}
\label{sec:cpt}

Before task-specific fine-tuning, we run a continued pre-training (CPT) stage that adapts the general-purpose JoyAI-Image-Edit foundation toward the try-on domain while preserving its broad multimodal prior. Concretely, we continue training the model on a balanced mixture that combines the general-purpose pretraining corpus and our try-on data at a $1{:}1$ ratio. Retaining an equal share of general-purpose data prevents the model from drifting away from its original generation and editing ability, while the try-on half injects domain knowledge---reference-item appearance, body-region placement, and wearing relations---so that the subsequent SFT stage starts from a substantially better initialization. The resulting CPT checkpoint serves as the base model for SFT (Section~\ref{sec:sft}).

\subsubsection{Supervised Fine-Tuning}
\label{sec:sft}

Building on the CPT checkpoint, the SFT stage specializes the model into an any-item try-on expert on the high-quality paired data produced by our data engine (Section~\ref{sec:data}). Unlike generic image editing, virtual try-on demands strong item-level fidelity and human-centric consistency: the model must reproduce the appearance of each reference item, adapt it to the target body pose and shape, keep the subject's identity and background intact, and synthesize physically plausible item--subject interactions. Accordingly, our SFT corpus spans diverse try-on scenarios---product-to-person try-on, person-to-person garment transfer, multi-item dressing, and accessories, shoes, and bags, as well as mask-free instruction-guided try-on---and supplies structured supervision from the reference items and subject to the final dressed result, so that the model learns the fine-grained correspondence between items and human regions.

\begin{table}[h]
\centering
\caption{Supervised fine-tuning hyperparameters.}
\label{tab:sft_hyperparams}
\begin{tabular*}{0.8\linewidth}{@{\extracolsep{\fill}} l c l c @{}}
\toprule
\textbf{Parameter} & \textbf{Value} & \textbf{Parameter} & \textbf{Value} \\
\midrule
Training Strategy & FSDP2 & Learning Rate & $5 \times 10^{-6}$ \\
Total Batch Size & 128 & LR Scheduler & Cosine \\
Micro Batch Size & 1 & Training Epochs & 3 \\
Mixed Precision & BF16 & Max Length & 8192 \\
\bottomrule
\end{tabular*}
\end{table}

To handle the inherent length heterogeneity of our training corpus, where single-reference and multi-reference try-on samples coexist, we adopt a dynamic sequence packing strategy that greedily packs multiple short samples into a single training sequence. Packed sub-sequences are kept mutually isolated through a block-diagonal mask in the variable-length attention interface, so that tokens from different samples never attend to one another, which reduces padding waste and improves training throughput. 

The model is trained with the conditional flow-matching objective of its MMDiT backbone~\cite{esser2024scaling}. Given a clean image latent $x_0$ (the VAE encoding of the target try-on result), a noise sample $\epsilon\sim\mathcal{N}(0,\mathbf{I})$, and a timestep $t\in[0,1]$, we form the interpolated latent $x_t=(1-t)\,x_0+t\,\epsilon$ and regress the predicted velocity field toward its target:

\begin{equation}
\mathcal{L}_{\mathrm{SFT}}
=
\mathbb{E}_{x_0,\epsilon,t}\Big[\big\lVert v_{\theta}(x_t, t, c) - (\epsilon - x_0)\big\rVert_2^2\Big],
\end{equation}

where the conditioning $c=(T, I_s, \mathcal{R})$ comprises the text instruction $T$, the subject image $I_s$, and the set of one or more reference item images $\mathcal{R}$. Overall, this SFT recipe equips the model with robust item transfer, multi-item composition, and mask-free try-on. The detailed training hyperparameters are summarized in Table~\ref{tab:sft_hyperparams}.

\subsubsection{Reinforcement Learning}
\label{sec:rl}

After SFT, we further optimize the model with reinforcement learning to push beyond the limits of likelihood-based training, targeting the qualities that matter most for try-on but are hard to capture with a pixel-level loss: item--subject consistency, item fidelity, identity and background preservation, and physical plausibility under complex poses and layering. We adopt the DiffusionNFT~\cite{zheng2025diffusionnft} RL paradigm and further draw on recent reward-based diffusion optimization methods, including Flow-GRPO~\cite{liu2025flow} and Dynamic-TreeRPO~\cite{fu2025dynamic}. The model is optimized with a hybrid reward that combines a rubric-guided VLM judge with an in-house try-on reward model.

%We adopt the DiffusionNFT~\cite{zheng2025diffusionnft} RL paradigm and drive it with a hybrid reward that combines a rubric-guided VLM judge with an in-house try-on reward model.

\noindent\textbf{RL data curation and balancing.} The quality and coverage of the RL prompt set directly shape what the policy learns to optimize, so we curate a dedicated RL training set through a combination of human annotation and model-based filtering, keeping only high-quality samples whose scenarios are balanced across several axes. Concretely, we balance the data along (i) the \emph{instruction} (\eg, single- vs.\ multi-item wearing commands, with or without an additional editing directive), (ii) the \emph{reference type} (clean product shots vs.\ worn-on photos), (iii) the \emph{scene complexity} (\eg, background clutter, pose difficulty, and occlusion), and (iv) the \emph{number of reference conditions}. This balanced distribution prevents the policy from overfitting to head categories or easy cases and ensures that the reward signal is exercised across the full range of try-on scenarios.

\noindent\textbf{Hybrid reward design.} In reinforcement learning, the model's behavior is ultimately shaped by its reward signal, which must evaluate generated outputs across multiple complementary dimensions: \emph{item fidelity} (faithfully reproducing texture, pattern, and logos for each reference item), \emph{identity and background preservation} (keeping the subject's face, body shape, and scene intact), \emph{instruction following} (correctly placing all items and executing any editing directives), and \emph{physical/structural plausibility} (avoiding anatomical errors and impossible garment intersections). Rather than relying on a single judge, we adopt a \emph{hybrid reward} that fuses two complementary sources: (i) an \textbf{in-house reward model} trained specifically for virtual try-on, and (ii) a \textbf{rubric-guided multimodal judge} powered by Gemini~3.1~Pro~\cite{gemini31pro}. We detail the two sources below and then describe how they are combined.

\noindent\textbf{In-house reward model.} We train an in-house reward model tailored to virtual try-on, which provides the optimization signal for policy-model RL and evaluates generated results along three complementary dimensions, including item consistency, identity (face) consistency, and overall visual quality. The model is a 8B multimodal VLM built upon Qwen3-VL-Instruct~\cite{qwen3vl2025} and fully fine-tuned on virtual try-on preference data. To explicitly capture these task-specific quality factors, we equip the backbone with three independent reward heads, one for each dimension. The final reward of this model is obtained by averaging the scalar scores from the three heads, enabling a comprehensive evaluation of candidate generations under the same try-on condition.

The training data of this in-house reward model is organized in a pairwise preference format and consists of about 100k human-annotated comparison pairs. Each pair contains two candidate try-on results generated by different models under the same subject, reference item, and wearing instruction. Human annotators provide scores for each candidate along three dimensions, including item consistency, identity (face) consistency, and overall visual quality. During training, we determine the preferred and dispreferred samples according to the annotated scores for each dimension, and optimize the model with an uncertainty-aware pairwise ranking loss \cite{ma2025hpsv3, wu2025editreward}. Specifically, the model predicts both the reward mean and the associated uncertainty variance for each candidate, while the pairwise ranking objective maximizes the probability that the preferred sample outperforms the dispreferred one. Unlike standard pairwise ranking supervision with deterministic preference scores, our formulation jointly models reward prediction and its uncertainty. This design explicitly accounts for subjectivity, near-tie comparisons, and potential noise in preference labels, thereby yielding more stable reward signals for fine-grained virtual try-on evaluation.

Since the reward model is directly trained on virtual try-on preference data and explicitly captures the three key dimensions, it can identify task-critical failure modes in generated results with more fine-grained sensitivity, such as missing item textures, silhouette distortion, identity drift, and abnormal human poses. Consequently, it provides more accurate and task-relevant reward signals for policy-model training.

\noindent\textbf{Rubric-guided Gemini reward.} In parallel, we query Gemini~3.1~Pro~\cite{gemini31pro} as a multimodal judge for rubric-based RL. Given the reference images and the textual instruction as input, the judge scores the generated try-on results along two top-level aspects---\emph{consistency} and \emph{visual quality}---each decomposed into three explicit, checkable criteria that are individually rated on a $0$--$10$ scale against anchored level descriptions:
\begin{itemize}
\item \emph{Consistency}: (i) \textbf{instruction consistency}---whether the changes requested by the instruction are applied and, equally important, whether nothing the instruction did \emph{not} mention is altered; (ii) \textbf{identity consistency}---faithful preservation of the subject's face; and (iii) \textbf{item consistency}---faithful reproduction of each reference item's texture, pattern, and structure.
\item \emph{Visual quality}: (i) the \textbf{naturalness and plausibility} of how the items are worn; (ii) the \textbf{aesthetic quality} of the overall image; and (iii) the \textbf{absence of artifacts}.
\end{itemize}
Decomposing each aspect into anchored criteria substantially improves the consistency and discriminativeness of the judgments compared with free-form prompting. To turn the six criteria into a single judge score, we first aggregate the three criteria \emph{within} each aspect using the harmonic mean, and then combine the two aspect-level scores using the geometric mean. Both choices penalize imbalance: a result cannot earn a high reward by excelling on one criterion while failing another, which curbs the per-dimension reward hacking that VLM-as-reward judges are otherwise prone to (\eg, copying the reference item perfectly while ignoring the instruction, or producing an aesthetically pleasing image that violates identity).

\noindent\textbf{Reward aggregation.} The two reward sources are complementary: the rubric-guided Gemini judge contributes a broad, instruction-level assessment with strong open-world generalization, while the in-house reward model contributes task-specific, fine-grained consistency cues distilled from our own try-on preference data. We fuse them into the single scalar reward used for policy optimization as a weighted average of the Gemini reward and the in-house try-on reward, so that the two signals jointly supervise both the holistic, instruction-level quality and the fine-grained consistency of each generation, and combining them mitigates the reward hacking and blind spots that either source would exhibit on its own.

\noindent\textbf{Policy optimization.} Built on the SFT checkpoint, we optimize the policy with DiffusionNFT~\cite{zheng2025diffusionnft}. For each try-on condition we draw a group of generations and compute a group-relative reward, normalizing each sample's reward against its group so that the objective favors higher-reward generations over lower-reward ones within the same group~\cite{shao2024deepseekmath}. This RL stage improves try-on consistency and stability, particularly under challenging poses, occlusions, and dense multi-item combinations.

\subsection{Inference and Prompt Design}
\label{sec:prompt}

At inference time, Oxygen-TryOn is driven by the textual instruction $T$ together with the subject image $I_s$ and the reference set $\mathcal{R}$. Because the model is understanding-driven, the wording of $T$ directly governs how the references are interpreted, associated with body regions, and composed. We therefore adopt a lightweight but systematic prompt design that makes the model's behavior predictable and exposes its full capability without any change to the network or weights.

\noindent\textbf{Structured instruction template.} We express $T$ in a structured form that separates the \emph{wearing intent} from an \emph{optional editing directive}. The wearing part enumerates each reference item and, when necessary, its target body region or layering order (\eg, ``dress the person in Picture~1 as the top and Picture~2 as the shoes''), so that the MLLM can establish an unambiguous item-to-region mapping for both single- and multi-item composition. The optional editing part appends general edits to be applied within the same generation pass (\eg, ``\dots\ and change the pose to a three-quarter view'').
In practice, we instantiate the instruction with the following slots:
\texttt{[garment\textunderscore image]}, \texttt{[subject\textunderscore image]}, \texttt{[item\textunderscore type]}, and the optional \texttt{[edit\textunderscore directive]}. Here, \texttt{[garment\textunderscore image]} denotes the source image that provides the garment or accessory to be transferred, \texttt{[subject\textunderscore image]} denotes the subject image to be edited, and \texttt{[item\textunderscore type]} specifies the semantic category of the transferred item, such as top, pants, dress, shoes, bag, or accessory. For multi-item try-on, the same slots are instantiated repeatedly for each reference item, and the image index in the instruction is used to bind each item to its corresponding source image. For single-item try-on, the template is:
``Put the \texttt{[item\textunderscore type]} from Picture 1 \texttt{[garment\textunderscore image]} onto the subject in Picture 2 \texttt{[subject\textunderscore image]}, keeping the subject's posture, face, and body shape unchanged.'' For multi-item try-on, the template enumerates all references: ``Using the subject in Picture 1 \texttt{[subject\textunderscore image]} as a base, modify the outfit by dressing the subject in the following ways: put the \texttt{[item\textunderscore type]} from Picture 2 on the subject, replacing the original \texttt{[item\textunderscore type]}; put the \texttt{[item\textunderscore type]} from Picture 3 on the subject, replacing the original \texttt{[item\textunderscore type]}; \ldots Keep the subject's facial features, hairstyle, body shape, posture, background, and lighting unchanged from Picture 1.'' For editing-augmented inference, we append the user-specified edit directive to the same instruction, e.g., ``\ldots, and remove the original jacket'' or ``\ldots, while keeping the handbag unchanged.'' If the editing directive explicitly modifies an attribute that is otherwise listed as preserved, the editing directive takes precedence; all unspecified attributes remain unchanged. When the prompt enhancer (PE) is enabled, an MLLM first predicts the task type and image roles, then fills these slots and outputs a canonical rewritten prompt.

\noindent\textbf{Reference ordering.} For multi-item try-on, the instruction disambiguates the references by their natural-language position (\eg, ``Picture~1'', ``Picture~2'') rather than through any special tag or identifier token (Section~\ref{sec:multiref}). This lets a single instruction control \emph{which} item goes \emph{where}, rather than relying on reference order alone. Nonetheless, to match the convention of our training data, we recommend ordering the references with product/garment shots first and any worn-on subject reference last, which best fits the training distribution and yields the most stable placement.

% \noindent\textbf{Training--inference consistency.} To avoid a train--test prompt gap, the instructions used at inference are drawn from the same instruction distribution as those synthesized for training by our data engine (Section~\ref{sec:data}). This synthesis pipeline proceeds in two steps. We first generate template-based instructions from structured annotations of image roles, item categories, reference indices, and edit operations. We then use an MLLM rewriter to paraphrase these instructions into natural user-like commands while preserving the source-image indices, item categories, target regions, and preservation constraints, so that training prompts cover the concise wearing commands, detailed multi-item descriptions, and edit-augmented requests that users actually employ at test time. At inference, PE applies the same rewriting policy to map free-form user prompts back to this canonical distribution. By aligning the two sides, this design reduces ambiguity in image roles and item-region associations---particularly for multi-reference try-on, accessories, and shoes---and keeps the model stable under the natural paraphrases encountered in real user inputs.

%% file: sections/experiments.tex
\section{Experiments}
\label{sec:experiments}

\subsection{Evaluation Benchmark}
\label{sec:benchmark}

To evaluate Oxygen-TryOn comprehensively, we use four complementary benchmarks that progress from constrained academic settings to unconstrained real-world conditions.

\noindent\textbf{Standard academic benchmarks.} DressCode~\cite{morelli2022dresscode} and VITON-HD~\cite{choi2021vitonhd} are the de facto benchmarks for image-based try-on. They consist primarily of simple scenes, such as clean product shots of a garment paired with a studio model image against a plain background, and thus mainly probe basic garment-transfer fidelity.
On these datasets, we report conventional reconstruction/perceptual metrics, where lower FID/KID/LPIPS and higher SSIM indicate better performance.

\noindent\textbf{TStars-VTON.} TStars-VTON~\cite{chen2026tstars} is a recent public benchmark covering rich and diverse scenarios, ranging from single-reference to multi-reference try-on.
For TStars-VTON, we follow its VLM-driven evaluation protocol~\cite{chen2026tstars}, which decomposes try-on quality into four dimensions, each scored on a $1$--$10$ scale: \emph{Identity Consistency} (preservation of the person's face, pose, and body shape), \emph{Item Fidelity} (faithful reproduction of each reference item's texture, pattern, and structure), \emph{Background Preservation} (integrity of the original scene), and \emph{Physical \& Structural Logic} (anatomical correctness and plausible layering/occlusion). Following this protocol, we use Gemini~3.1~Pro as the judge. The \emph{Overall} score is computed as the geometric mean of the four dimensions, so that a model must achieve balanced excellence across all axes. 

\noindent\textbf{Oxygen-TryOn Bench.} To measure performance under realistic deployment conditions, we further construct Oxygen-TryOn Bench, an internal benchmark whose images are all collected from real-world scenes. It covers diverse items including garments, jewelry, shoes, and bags, and explicitly incorporates the challenges of practical use, such as multi-item layering, complex backgrounds, diverse human poses, and both full-body and half-body views. The reference items are provided either as product display images or as photos of a person already wearing the item, and a single sample may involve one or multiple items. According to the form of the reference, we divide the benchmark into two scenarios: \emph{Cloth-to-Model}, where the reference is a clean product-display image of the item (\eg, a flat-lay or mannequin shot) that must be transferred onto the subject, and \emph{Model-to-Model}, where the reference is a photo of one person already wearing the item and the goal is to re-dress a different subject in the same item. The two scenarios pose distinct challenges: Cloth-to-Model emphasizes faithfully reconstructing item details from a context-free product shot, whereas Model-to-Model additionally requires disentangling the item from the source person before re-applying it to the target. The benchmark contains $1{,}000$ test samples in total, evenly split between the two scenarios, with $500$ Cloth-to-Model samples and $500$ Model-to-Model samples.

For the Oxygen-TryOn Bench, we use GPT-5 as the judge and score each result on a $1$--$5$ scale along three deployment-oriented dimensions---\emph{Subject Consistency}, \emph{Item Consistency}, and \emph{Aesthetics}. We report an \emph{Overall} score, computed as the average of the three dimensions, together with a \emph{Usability Rate}, defined as the fraction of results whose three dimension scores are \emph{all} strictly above $3$ (\textit{i.e.}, directly shippable without manual rework); the product-display (Cloth-to-Model) and worn-on (Model-to-Model) splits are evaluated separately. We additionally conduct a human evaluation against the strongest proprietary competitors (Section~\ref{sec:human_eval}), scoring subject consistency, item consistency, aesthetics, and an aggregate overall score.

\subsection{Quantitative Results}
\label{sec:quant}
\providecommand{\reprostar}{\raisebox{0.55ex}{\scriptsize\hspace{-0.06em}\textnormal{*}}}

We compare Oxygen-TryOn against representative academic try-on models~\cite{choi2024idmvton,xu2025ootdiffusion,chong2024catvtonconcatenationneedvirtual,zhou2024learning,jiang2024fitdit,kim2024promptdresser,guo2025any2anytryon,chong2025fastfitacceleratingmultireferencevirtual} and strong general-purpose generation/editing models, including open-source~\cite{wu2025qwen,blackforest2025flux2klein} and proprietary systems~\cite{gptimage2_model_card,google2025nanobanana,bytedance2026seedream}. For the conventional DressCode and VITON-HD comparisons, the paired results are taken from the official FastFit paper~\cite{chong2025fastfitacceleratingmultireferencevirtual}, while the unpaired results are taken from the official TStars-Tryon1.0 paper~\cite{chen2026tstars}. Because we could not fully reproduce the originally reported numbers for several baselines, for these methods we list both their officially reported numbers and the results reproduced in our own environment for a fair comparison, where the reproduced rows are marked with \reprostar{}. TStars-Tryon1.0 is not open-sourced, so we are unable to reproduce it and only report its available official unpaired results. Results on conventional paired/unpaired benchmarks are reported in Table~\ref{tab:standard_results}; TStars-VTON single-item and multi-item results are reported in Table~\ref{tab:single_results} and Table~\ref{tab:multi_results}, respectively. Figure~\ref{fig:teaser_radar} summarizes the per-dimension TStars-VTON profiles at a glance.

\input{sections/radar}

\begin{table}[t]
\centering
\caption{\textbf{Quantitative results on DressCode and VITON-HD.} \emph{Paired} evaluation measures each generated try-on result against its paired ground-truth target image and reports FID, KID, SSIM, and LPIPS; \emph{unpaired} evaluation compares generated images with the real-image distribution without one-to-one ground truth and reports FID and KID. For each dataset, rows are grouped into \textit{Officially Reported} numbers, taken from the original papers, and results \textit{Re-evaluated in Our Environment} for methods whose reported numbers we could not reproduce exactly (marked with \reprostar{}). For a fair comparison, best and second-best results are computed only among the re-evaluated (\reprostar) methods and ours, shown in \textbf{bold} and \underline{underlined}; the officially reported scores are listed for reference only.}
\label{tab:standard_results}
{\footnotesize
\renewcommand{\arraystretch}{1.05}
\begin{tabular}{llcccccc}
\toprule
\multirow{2}{*}{Dataset} & \multirow{2}{*}{Method} & \multicolumn{4}{c}{Paired} & \multicolumn{2}{c}{Unpaired} \\
\cmidrule(lr){3-6}\cmidrule(lr){7-8}
& & FID $\downarrow$ & KID $\downarrow$ & SSIM $\uparrow$ & LPIPS $\downarrow$ & FID $\downarrow$ & KID $\downarrow$ \\
\midrule
\multirow{15}{*}{DressCode}
& \multicolumn{7}{l}{\cellcolor[gray]{0.95}\textit{Officially Reported}} \\
& IDM-VTON~\cite{choi2024idmvton} & 7.181 & 3.524 & 0.891 & 0.070 & 9.167 & 4.489 \\
& OOTDiffusion~\cite{xu2025ootdiffusion} & 6.975 & 2.014 & 0.873 & 0.077 & 8.121 & 2.886 \\
& FitDiT~\cite{jiang2024fitdit} & 5.571 & 1.901 & 0.899 & 0.058 & 4.805 & 0.712 \\
& CatVTON~\cite{chong2024catvtonconcatenationneedvirtual} & 3.710 & 1.010 & 0.909 & 0.062 & 5.872 & 1.606 \\
& Leffa~\cite{zhou2024learning} & 7.193 & 2.114 & 0.861 & 0.084 & 20.099 & 13.506 \\
& PromptDresser~\cite{kim2024promptdresser} & 9.563 & 4.795 & 0.858 & 0.104 & 10.618 & 4.978 \\
& Any2AnyTryon~\cite{guo2025any2anytryon} & 5.111 & 1.265 & 0.897 & 0.059 & 6.709 & 1.580 \\
& FastFit~\cite{chong2025fastfitacceleratingmultireferencevirtual} & 2.836 & 0.390 & 0.907 & 0.057 & 4.397 & 0.553 \\
& TStars-Tryon1.0~\cite{chen2026tstars} & -- & -- & -- & -- & 4.541 & 0.458 \\
\cmidrule(lr){2-8}
& \multicolumn{7}{l}{\cellcolor[gray]{0.95}\textit{Re-evaluated in Our Environment}} \\
& CatVTON\reprostar~\cite{chong2024catvtonconcatenationneedvirtual} & 5.031 & 1.789 & \underline{0.899} & 0.074 & 7.485 & 2.717 \\
& Any2AnyTryon\reprostar~\cite{guo2025any2anytryon} & 4.893 & 1.252 & 0.881 & 0.082 & 6.540 & 1.667 \\
& FastFit\reprostar~\cite{chong2025fastfitacceleratingmultireferencevirtual} & \underline{3.658} & \underline{0.747} & 0.889 & \underline{0.070} & \underline{5.357} & \underline{1.027} \\
\rowcolor{rowyellow}& \textbf{Oxygen-TryOn (Ours)} & \textbf{1.932} & \textbf{0.387} & \textbf{0.936} & \textbf{0.034} & \textbf{5.155} & \textbf{0.945} \\
\midrule
\multirow{15}{*}{VITON-HD}
& \multicolumn{7}{l}{\cellcolor[gray]{0.95}\textit{Officially Reported}} \\
& IDM-VTON~\cite{choi2024idmvton} & 6.112 & 1.112 & 0.866 & 0.074 & 9.249 & 1.267 \\
& OOTDiffusion~\cite{xu2025ootdiffusion} & 5.762 & 0.267 & 0.843 & 0.072 & 9.082 & 0.702 \\
& FitDiT~\cite{jiang2024fitdit} & 8.176 & 1.079 & 0.838 & 0.096 & 9.979 & 1.478 \\
& CatVTON~\cite{chong2024catvtonconcatenationneedvirtual} & 6.738 & 1.320 & 0.881 & 0.088 & 10.552 & 2.272 \\
& Leffa~\cite{zhou2024learning} & 5.667 & 0.692 & 0.857 & 0.076 & 10.446 & 2.640 \\
& PromptDresser~\cite{kim2024promptdresser} & 5.934 & 0.550 & 0.846 & 0.090 & 8.885 & 0.909 \\
& Any2AnyTryon~\cite{guo2025any2anytryon} & 11.195 & 2.806 & 0.799 & 0.194 & 9.981 & 3.496 \\
& FastFit~\cite{chong2025fastfitacceleratingmultireferencevirtual} & 5.620 & 0.505 & 0.885 & 0.078 & 8.629 & 0.665 \\
& TStars-Tryon1.0~\cite{chen2026tstars} & -- & -- & -- & -- & 8.485 & 0.528 \\
\cmidrule(lr){2-8}
& \multicolumn{7}{l}{\cellcolor[gray]{0.95}\textit{Re-evaluated in Our Environment}} \\
& CatVTON\reprostar~\cite{chong2024catvtonconcatenationneedvirtual} & 8.055 & 1.659 & \underline{0.888} & \underline{0.076} & 11.404 & 2.374 \\
& Any2AnyTryon\reprostar~\cite{guo2025any2anytryon} & 11.532 & 3.884 & 0.875 & 0.129 & 14.756 & 5.084 \\
& FastFit\reprostar~\cite{chong2025fastfitacceleratingmultireferencevirtual} & \underline{6.863} & \underline{0.949} & 0.865 & 0.096 & \underline{9.491} & \textbf{0.973} \\
\rowcolor{rowyellow}& \textbf{Oxygen-TryOn (Ours)} & \textbf{3.941} & \textbf{0.491} & \textbf{0.914} & \textbf{0.050} & \textbf{9.145} & \underline{1.171} \\
\bottomrule
\end{tabular}
}
\end{table}

\begin{table}[t]
\centering
\caption{\textbf{Quantitative results on TStars-VTON (single-item try-on).} We evaluate all models with the official TStars-VTON evaluation scripts. However, possibly due to changes in the official Gemini API, our reproduced scores differ substantially from those reported in the TStars-VTON paper; therefore, for a fair comparison, we re-evaluate the current scoring performance of all methods under the official scripts (re-evaluated rows are marked with \reprostar{}). Best and second-best results, computed only among the re-evaluated (\reprostar) scores and ours, are in \textbf{bold} and \underline{underlined}; the officially reported scores are listed for reference only.}
\label{tab:single_results}
{\footnotesize
\begin{tabular}{lccccc}
\toprule
Method & Overall\,$\uparrow$ & \makecell{Identity \\ Consist.}\,$\uparrow$ & \makecell{Item \\ Fidelity}\,$\uparrow$ & \makecell{Backgr. \\ Preserv.}\,$\uparrow$ & \makecell{Phys. \& \\ Struc. Logic}\,$\uparrow$ \\
\midrule
\rowcolor[gray]{0.95} \textit{Officially Reported} & & & & & \\
CatVTON~\cite{chong2024catvtonconcatenationneedvirtual} & 6.66 & 9.34 & 4.01 & 9.47 & 7.96 \\
Leffa~\cite{zhou2024learning} & 6.05 & 8.14 & 4.35 & 8.75 & 6.01 \\
FitDiT~\cite{jiang2024fitdit} & 5.15 & 6.77 & 4.71 & 8.03 & 3.88 \\
FastFit~\cite{chong2025fastfitacceleratingmultireferencevirtual} & 6.45 & 9.13 & 4.67 & 8.34 & 6.55 \\
Qwen-Image-Edit-2511~\cite{wu2025qwen} & 8.12 & 9.21 & 6.79 & 9.17 & 8.87 \\
FLUX.2-dev~\cite{blackforest2025flux2klein} & 8.76 & 9.42 & 7.92 & 9.64 & 8.96 \\
FLUX.2-klein~\cite{blackforest2025flux2klein} & 8.80 & 9.44 & 8.18 & 9.50 & 8.90 \\
FireRed-Image-Edit1.1~\cite{team2026firered} & 8.86 & 9.61 & 7.80 & 9.78 & 9.07 \\
GPT-Image-1.5~\cite{gptimage15_model_card} & 8.89 & 9.38 & 8.56 & 9.08 & 9.22 \\
GPT-Image-2~\cite{gptimage2_model_card} & 9.20 & 9.60 & 8.79 & 9.59 & 9.26 \\
Nano Banana Pro~\cite{google2025nanobanana} & 9.23 & 9.86 & 8.60 & 9.82 & 9.19 \\
Seedream5 Lite~\cite{bytedance2026seedream} & 9.30 & 9.85 & 8.64 & 9.81 & 9.34 \\
TStars-Tryon1.0~\cite{chen2026tstars} &9.37 &9.89 &8.83 &9.86 &9.24\\
\midrule
\rowcolor[gray]{0.95} \textit{Re-evaluated under Official Scripts} & & & & & \\
FastFit\reprostar~\cite{chong2025fastfitacceleratingmultireferencevirtual} & 5.07 & 8.05 & 3.72 & 7.46 & 4.38 \\
Qwen-Image-Edit-2511\reprostar~\cite{wu2025qwen} & 6.10 & 7.77 & 6.04 & 6.31 & 7.14 \\
FLUX.2-dev\reprostar~\cite{blackforest2025flux2klein} & 7.76 & 8.50 & 6.55 & 8.50 & 7.66 \\
FireRed-Image-Edit1.1\reprostar~\cite{team2026firered} & 7.96 & 9.18 & 7.07 & 9.18 & 8.27 \\
GPT-Image-2\reprostar~\cite{gptimage2_model_card} & 8.34 & 9.50 & 8.51 & 8.46 & 8.21 \\
Nano Banana Pro\reprostar~\cite{google2025nanobanana} & 8.04 & 9.15 & \underline{8.65} & 7.88 & 8.13 \\
Seedream5 Lite\reprostar~\cite{bytedance2026seedream} & \underline{8.77} & \underline{9.67} & 8.61 & \underline{9.19} & \underline{8.48} \\
\midrule
\rowcolor{rowyellow} \textbf{Oxygen-TryOn (Ours)} & \textbf{9.36} & \textbf{9.82} & \textbf{9.10} & \textbf{9.69} & \textbf{9.29} \\
\bottomrule
\end{tabular}
}
\end{table}

\begin{table}[t]
\centering
\caption{\textbf{Quantitative results on TStars-VTON (multi-item try-on), evaluated on samples with 2--4 reference conditions.} As in Table~\ref{tab:single_results}, all methods are evaluated with the official TStars-VTON scripts; rows marked with \reprostar{} are re-evaluated by us under these scripts. Best and second-best results, computed only among the re-evaluated (\reprostar) methods and ours, are in \textbf{bold} and \underline{underlined}.}
\label{tab:multi_results}
{\footnotesize
\begin{tabular}{lccccc}
\toprule
Method & Overall\,$\uparrow$ & \makecell{Identity \\ Consist.}\,$\uparrow$ & \makecell{Item \\ Fidelity}\,$\uparrow$ & \makecell{Backgr. \\ Preserv.}\,$\uparrow$ & \makecell{Phys. \& \\ Struc. Logic}\,$\uparrow$ \\
\midrule
\rowcolor[gray]{0.95} \textit{Re-evaluated under Official Scripts} & & & & & \\
Qwen-Image-Edit-2511\reprostar~\cite{wu2025qwen} & 2.35 & 2.50 & 3.35 & 1.73 & 4.70 \\
FLUX.2-dev\reprostar~\cite{blackforest2025flux2klein} & 4.59 & 7.73 & 3.12 & 7.06 & 4.04 \\
FireRed-Image-Edit1.1\reprostar~\cite{team2026firered} & 5.51 & 5.90 & 6.11 & 5.57 & 7.54 \\
GPT-Image-2\reprostar~\cite{gptimage2_model_card} & 7.90 & \underline{9.03} & 8.33 & 8.19 & 7.71 \\
Nano Banana Pro\reprostar~\cite{google2025nanobanana} & 7.69 & 8.62 & \textbf{8.50} & 7.58 & 7.85 \\
Seedream5 Lite\reprostar~\cite{bytedance2026seedream} & \underline{8.19} & 8.78 & \underline{8.49} & \underline{8.70} & \underline{8.09} \\
\midrule
\rowcolor{rowyellow} \textbf{Oxygen-TryOn (Ours)} & \textbf{8.41} & \textbf{9.07} & 8.03 & \textbf{9.00} & \textbf{8.57} \\
\bottomrule
\end{tabular}
}
\end{table}

\paragraph{Standard academic benchmarks.} Table~\ref{tab:standard_results} reports paired and unpaired results on DressCode and VITON-HD. For a fair comparison, we highlight the best and second-best results only among the methods re-evaluated in our own environment (marked with \reprostar) and Oxygen-TryOn, and list officially reported numbers for reference only. On the paired protocol, which directly measures reconstruction fidelity against the ground-truth target, Oxygen-TryOn ranks first on every metric on both datasets---FID ($1.932$), KID ($0.387$), SSIM ($0.936$), and LPIPS ($0.034$) on DressCode, and FID ($3.941$), KID ($0.491$), SSIM ($0.914$), and LPIPS ($0.050$) on VITON-HD---with clear margins over the strongest re-evaluated baseline (\eg, FastFit\reprostar\ at DressCode paired FID $3.658$), indicating that our unified model reproduces item appearance and subject attributes more faithfully. On the unpaired protocol, which instead measures distributional similarity to the real-image set, Oxygen-TryOn again leads the re-evaluated comparison on DressCode (FID $5.155$ / KID $0.945$) and on VITON-HD FID ($9.145$), ranking a close second only on VITON-HD unpaired KID ($1.171$, behind FastFit\reprostar's $0.973$). Some specialized methods report even lower unpaired FID/KID in their original papers; we attribute this to their close alignment with each benchmark's narrow studio-photo distribution, whereas Oxygen-TryOn is optimized for faithful, instruction-driven try-on across far more diverse, in-the-wild scenarios.

\paragraph{Single-item try-on.} On the TStars-VTON single-item setting (Table~\ref{tab:single_results}), specialized academic models preserve identity and background well, e.g., CatVTON reaches $9.34$ on identity and $9.47$ on background, but collapse on item fidelity ($4.01$ for CatVTON, $4.67$ for FastFit), reflecting their difficulty in reproducing fine item detail in complex scenes. Re-evaluating the general-purpose and proprietary systems under the official scripts yields markedly lower scores than their originally reported numbers, e.g., GPT-Image-2 drops from $9.20$ to $8.34$ overall and Seedream5 Lite from $9.30$ to $8.77$, confirming the API-induced discrepancy noted in the table caption. Comparing fairly against these re-evaluated scores, Oxygen-TryOn ranks first on all five dimensions, with an overall score of $9.36$ versus $8.77$ for the next-best re-evaluated system and a particularly large lead on item fidelity ($9.10$ vs.\ $8.65$).

\paragraph{Multi-item try-on.} The difficulty escalates sharply in multi-item scenarios, where a model must coordinate several references and resolve layering and occlusion (Table~\ref{tab:multi_results}). Because our foundation model is pretrained for at most four reference images, we focus this evaluation on samples with up to four reference conditions, where Oxygen-TryOn operates within its primary capability range. In this setting, Oxygen-TryOn leads the re-evaluated comparison on four of five dimensions, including overall quality ($8.41$ vs.\ $8.19$ for the next-best Seedream5 Lite), identity ($9.07$), background ($9.00$), and physical plausibility ($8.57$); only item fidelity ($8.03$) trails the strongest proprietary systems (Nano Banana Pro $8.50$, Seedream5 Lite $8.49$, GPT-Image-2 $8.33$); we attribute this to their stronger texture reproduction and to the greater difficulty of preserving fine per-item detail under multi-item layering, a gap bounded by the capacity of our pretrained base model that we aim to further strengthen. Coordinating five or more references under complex layering likewise remains challenging and is a primary target for future work.

\paragraph{In-the-wild try-on.} Beyond the academic and TStars-VTON benchmarks, we assess real-world readiness on our Oxygen-TryOn Bench (Section~\ref{sec:benchmark}), which contains $1{,}000$ samples evenly split between product-display (\emph{Cloth-to-Model}, $500$ samples) and worn-on (\emph{Model-to-Model}, $500$ samples) references. As reported in Table~\ref{tab:inthewild}, Oxygen-TryOn ranks first on $8$ of the $10$ metrics across the two splits, leading on subject consistency, item consistency, the overall score, and the usability rate under both reference styles. Most notably, it raises the usability rate---the fraction of directly shippable results---to $86.79\%$ on the Cloth-to-Model split and $85.43\%$ on the harder Model-to-Model split, well above the strongest proprietary competitor GPT-Image-2 ($80.35\%$ and $77.58\%$) and far above the best open-source baselines (at most $67.48\%$ on the Cloth-to-Model split, from FLUX.2-dev, and $68.43\%$ on the Model-to-Model split, from FLUX.2-dev). The only dimension on which Oxygen-TryOn does not rank first is aesthetics, where it is a close second on the Cloth-to-Model split (behind GPT-Image-2, $3.736$ vs.\ $3.768$) and third on the Model-to-Model split (behind GPT-Image-2 and Seedream5 Lite); even there, its absolute scores stay high and within a narrow margin of the best. These results indicate that, under realistic deployment conditions, Oxygen-TryOn yields substantially more usable try-on results than both open-source and proprietary systems.

\begin{table}[t]
\centering
\caption{\textbf{Quantitative results on Oxygen-TryOn Bench.} We evaluate two reference styles separately: \emph{Cloth-to-Model} (product-display references, $500$ samples) and \emph{Model-to-Model} (worn-on references, $500$ samples). For each style we report Subject Consistency, Item Consistency, and Aesthetics on a $1$--$5$ scale, an aggregate \emph{Overall} score, and the \emph{Usability Rate} (the fraction of results whose three dimension scores are all strictly above $3$). Higher is better for all columns ($\uparrow$). Best and second-best results in each column are in \textbf{bold} and \underline{underlined}.}
\label{tab:inthewild}
{\footnotesize
\setlength{\tabcolsep}{3.2pt}
\renewcommand{\arraystretch}{1.1}
\resizebox{\textwidth}{!}{
\begin{tabular}{lcccccccccc}
\toprule
\multirow{2}{*}{Method} & \multicolumn{5}{c}{Cloth-to-Model (product-display)} & \multicolumn{5}{c}{Model-to-Model (worn-on)} \\
\cmidrule(lr){2-6}\cmidrule(lr){7-11}
& \makecell{Subject\\Consist.}\,$\uparrow$ & \makecell{Item\\Consist.}\,$\uparrow$ & Aesthetics\,$\uparrow$ & Overall\,$\uparrow$ & \makecell{Usability\\Rate (\%)}\,$\uparrow$ & \makecell{Subject\\Consist.}\,$\uparrow$ & \makecell{Item\\Consist.}\,$\uparrow$ & Aesthetics\,$\uparrow$ & Overall\,$\uparrow$ & \makecell{Usability\\Rate (\%)}\,$\uparrow$ \\
\midrule
\rowcolor[gray]{0.95} \multicolumn{11}{l}{\textit{Open-Source}} \\
FastFit~\cite{chong2025fastfitacceleratingmultireferencevirtual} & 3.650 & 2.173 & 3.257 & 3.027 & 34.96 & 3.621 & 1.963 & 3.185 & 2.923 & 25.51 \\
Qwen-Image-Edit-2511~\cite{wu2025qwen} & 3.287 & 2.627 & 3.667 & 3.194 & 53.18 & 3.278 & 2.660 & 3.493 & 3.144 & 55.90 \\
FireRed-Image-Edit1.1~\cite{team2026firered} & 3.673 & 2.637 & 3.685 & 3.332 & 64.28 & 3.333 & 2.742 & 3.670 & 3.248 & 63.31 \\
FLUX.2-dev~\cite{blackforest2025flux2klein} & 3.553 & 2.913 & 3.636 & 3.367 & 67.48 & 3.389 & 2.829 & 3.700 & 3.306 & 68.43 \\
\midrule
\rowcolor[gray]{0.95} \multicolumn{11}{l}{\textit{Closed-Source}} \\
Nano Banana Pro~\cite{google2025nanobanana} & 3.828 & 2.763 & 3.702 & 3.431 & 70.16 & 3.529 & 3.075 & 3.680 & 3.539 & 73.18 \\
Seedream5 Lite~\cite{bytedance2026seedream} & 3.820 & \underline{2.974} & 3.728 & 3.507 & 77.67 & \underline{3.631} & \underline{3.184} & \underline{3.802} & 3.428 & 74.70 \\
GPT-Image-2~\cite{gptimage2_model_card} & \underline{3.864} & 2.964 & \textbf{3.768} & \underline{3.532} & \underline{80.35} & 3.621 & 3.132 & \textbf{3.872} & \underline{3.542} & \underline{77.58} \\
\midrule
\rowcolor{rowyellow} \textbf{Oxygen-TryOn (Ours)} & \textbf{3.945} & \textbf{3.045} & \underline{3.736} & \textbf{3.575} & \textbf{86.79} & \textbf{3.870} & \textbf{3.340} & 3.757 & \textbf{3.656} & \textbf{85.43} \\
\bottomrule
\end{tabular}
}
}
\end{table}

\paragraph{Effect of the training recipe.}
\label{sec:effect_rl}
To isolate the contribution of each training stage, we evaluate three checkpoints of Oxygen-TryOn---the continued-pretraining base (\emph{CPT}), the supervised fine-tuned model (\emph{SFT}), and the final reinforcement-learning model (\emph{RL})---on our Oxygen-TryOn Bench under both reference styles (Table~\ref{tab:sft_vs_rl_tryon}). SFT delivers the largest single jump: relative to the CPT base, it raises the usability rate from $67.95\%$ to $85.95\%$ on the Cloth-to-Model split and from $70.18\%$ to $80.44\%$ on the Model-to-Model split, together with large gains in subject consistency (e.g., $3.228\rightarrow3.861$ on Model-to-Model) and the overall score, confirming that try-on-specific SFT installs the core wearing capability. RL then provides consistent further refinement on top of SFT: it improves item consistency ($2.984\rightarrow3.045$ and $3.286\rightarrow3.340$), aesthetics, and the overall score on both splits, and most notably lifts the harder Model-to-Model usability rate by a further $4.99$ points to $85.43\%$. The only minor regression is a marginal drop in Cloth-to-Model subject consistency ($3.963\rightarrow3.945$), which is outweighed by gains on every other dimension. Overall, the two stages play complementary roles: SFT establishes broad try-on competence, while RL sharpens fine-grained consistency and deployment-ready usability.

\begin{table}[tbhp]
\centering
\caption{\textbf{Effect of the training recipe on Oxygen-TryOn Bench.} We compare three stages of Oxygen-TryOn---the continued-pretraining base (CPT), the supervised fine-tuned model (SFT), and the final reinforcement-learning model (RL)---under the two reference styles: \emph{Cloth-to-Model} (product-display references) and \emph{Model-to-Model} (worn-on references). For each style we report Subject Consistency, Item Consistency, and Aesthetics on a $1$--$5$ scale, an aggregate \emph{Overall} score, and the \emph{Usability Rate} (the fraction of results whose three dimension scores are all strictly above $3$). Higher is better for all columns ($\uparrow$); the best result in each column is in \textbf{bold}.}
\label{tab:sft_vs_rl_tryon}
{\scriptsize
\setlength{\tabcolsep}{3.2pt}
\renewcommand{\arraystretch}{1.1}
\resizebox{\textwidth}{!}{
\begin{tabular}{lcccccccccc}
\toprule
\multirow{2}{*}{Model} & \multicolumn{5}{c}{Cloth-to-Model (product-display)} & \multicolumn{5}{c}{Model-to-Model (worn-on)} \\
\cmidrule(lr){2-6}\cmidrule(lr){7-11}
& \makecell{Subject\\Consist.}\,$\uparrow$ & \makecell{Item\\Consist.}\,$\uparrow$ & Aesthetics\,$\uparrow$ & Overall\,$\uparrow$ & \makecell{Usability\\Rate (\%)}\,$\uparrow$ & \makecell{Subject\\Consist.}\,$\uparrow$ & \makecell{Item\\Consist.}\,$\uparrow$ & Aesthetics\,$\uparrow$ & Overall\,$\uparrow$ & \makecell{Usability\\Rate (\%)}\,$\uparrow$ \\
\midrule
Oxygen-TryOn (CPT) & 3.731 & 2.888 & 3.685 & 3.435 & 67.95 & 3.228 & 2.731 & 3.703 & 3.220 & 70.18 \\
Oxygen-TryOn (SFT) & \textbf{3.963} & 2.984 & 3.711 & 3.553 & 85.95 & 3.861 & 3.286 & 3.657 & 3.601 & 80.44 \\
\rowcolor{rowyellow} Oxygen-TryOn (RL) & 3.945 & \textbf{3.045} & \textbf{3.736} & \textbf{3.575} & \textbf{86.79} & \textbf{3.870} & \textbf{3.340} & \textbf{3.757} & \textbf{3.656} & \textbf{85.43} \\
\bottomrule
\end{tabular}
}
}
\end{table}

\subsection{Human Evaluation}
\label{sec:human_eval}

To complement automated metrics, we conduct a human study comparing Oxygen-TryOn against the two strongest proprietary competitors, Nano Banana Pro~\cite{google2025nanobanana} and GPT-Image-2~\cite{gptimage2_model_card}. Trained human annotators rate each generated try-on on a $1$--$5$ scale along three dimensions: \emph{subject consistency} (faithful preservation of the subject's identity, body shape, and pose), \emph{item consistency} (accurate reproduction of each reference item's texture, pattern, and structure), and \emph{aesthetics} (overall realism and visual quality, free of artifacts). We further report an \emph{Overall} score, computed as a weighted average of the three dimensions, and higher is better for all four columns. The study is conducted on a shared set of roughly $1{,}000$ samples ($985$ in total) that all compared methods can process. As summarized in Table~\ref{tab:human_eval}, Oxygen-TryOn attains the best subject consistency ($3.7492$) and the best overall score ($3.5502$), the latter narrowly ahead of GPT-Image-2 ($3.5375$), while GPT-Image-2 holds a slight edge on item consistency ($3.2951$ vs.\ $3.2250$) and aesthetics ($3.8240$ vs.\ $3.8025$); Nano Banana Pro trails on every dimension. Overall, Oxygen-TryOn matches or surpasses the strongest proprietary systems in human preference, with its clearest advantage in preserving the subject. To mitigate individual subjectivity, each sample is independently assessed by nine trained annotators, with the three evaluation dimensions---subject consistency, item consistency, and aesthetics---each rated by three annotators, and every per-dimension score is averaged over its three raters to obtain the final rating.

\begin{table}[t]
\centering
\caption{\textbf{Human evaluation against strong proprietary systems.} Trained annotators rate each result on a $1$--$5$ scale for subject consistency, item consistency, and aesthetics; \emph{Overall} is a weighted average of the three dimensions. Higher is better for all columns ($\uparrow$). Best and second-best results are in \textbf{bold} and \underline{underlined}.}
\label{tab:human_eval}
{\footnotesize
\begin{tabular}{lcccc}
\toprule
Method & \makecell{Subject\\Consist.}\,$\uparrow$ & \makecell{Item\\Consist.}\,$\uparrow$ & Aesthetics\,$\uparrow$ & Overall\,$\uparrow$ \\
\midrule
Nano Banana Pro~\cite{google2025nanobanana} & 3.1959 & 2.9648 & 3.7174 & 3.2078 \\
GPT-Image-2~\cite{gptimage2_model_card} & \underline{3.6365} & \textbf{3.2951} & \textbf{3.8240} & \underline{3.5375} \\
\midrule
\rowcolor{rowyellow} \textbf{Oxygen-TryOn (Ours)} & \textbf{3.7492} & \underline{3.2250} & \underline{3.8025} & \textbf{3.5502} \\
\bottomrule
\end{tabular}
}
\end{table}

\subsection{Qualitative Results}
\label{sec:qualitative}

\begin{figure}[H]
    \centering
    \includegraphics[width=0.95\linewidth]{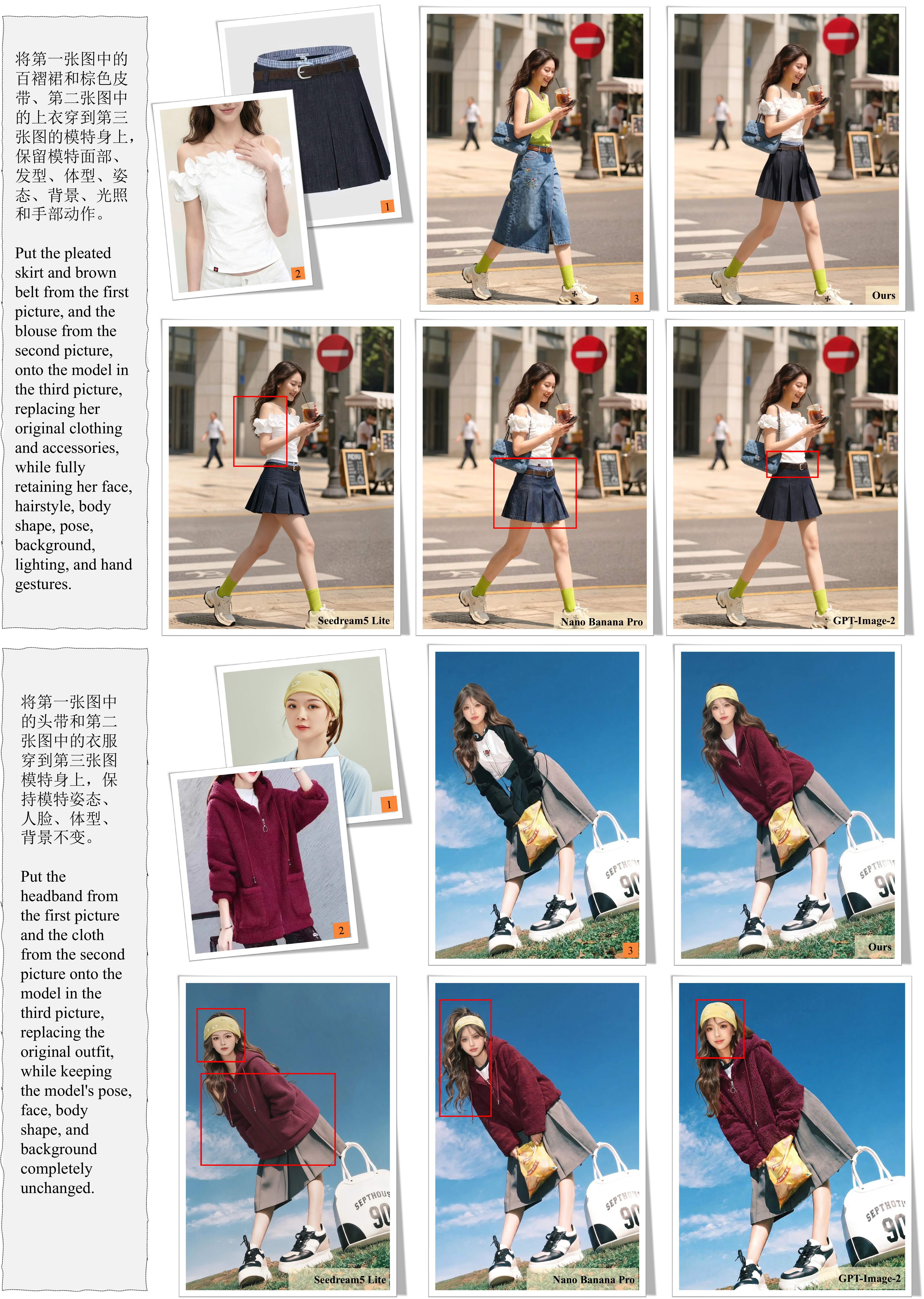}
    \caption{\textbf{Qualitative comparison with state-of-the-art proprietary systems.} \textcolor{red}{Red} boxes highlight local defects of the competing methods, such as distorted garments, mishandled belts and waistlines, and identity drift around newly added items.}
    \label{fig:cmp_multi}
\end{figure}

\begin{figure}[H]
    \centering
    \includegraphics[width=\linewidth]{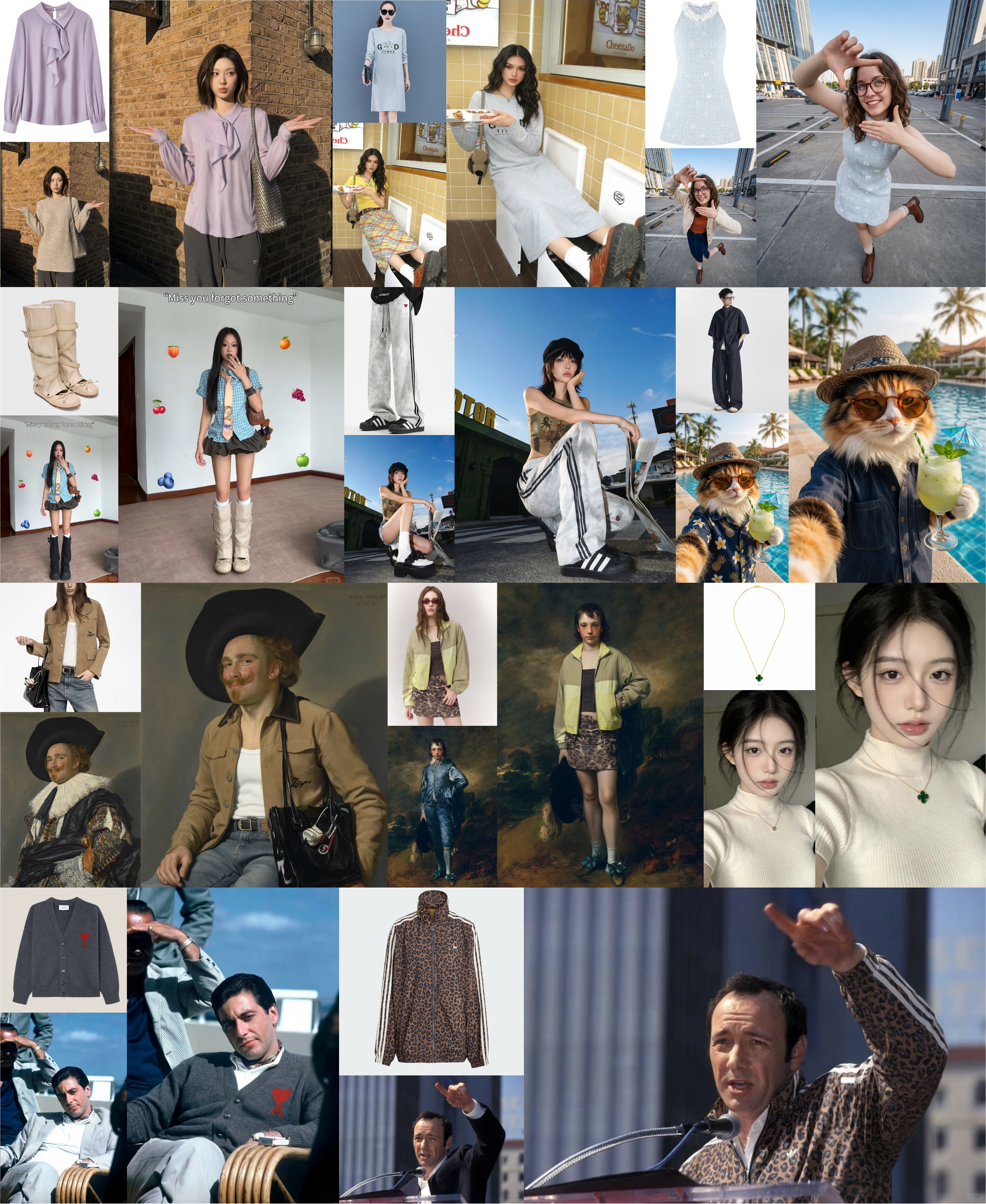}
    \caption{\textbf{Qualitative gallery of Oxygen-TryOn on single-item try-on.} Given a single reference---product shots or in-the-wild worn-on photos of garments, shoes, bags, and accessories---Oxygen-TryOn dresses diverse subjects while preserving item appearance (texture, pattern, and logos) and the subject's identity, body shape, and background across varied poses and full-/half-body views.}
\label{fig:gallery_single}
\end{figure}

\begin{figure}[H]
    \centering
    \includegraphics[width=\linewidth]{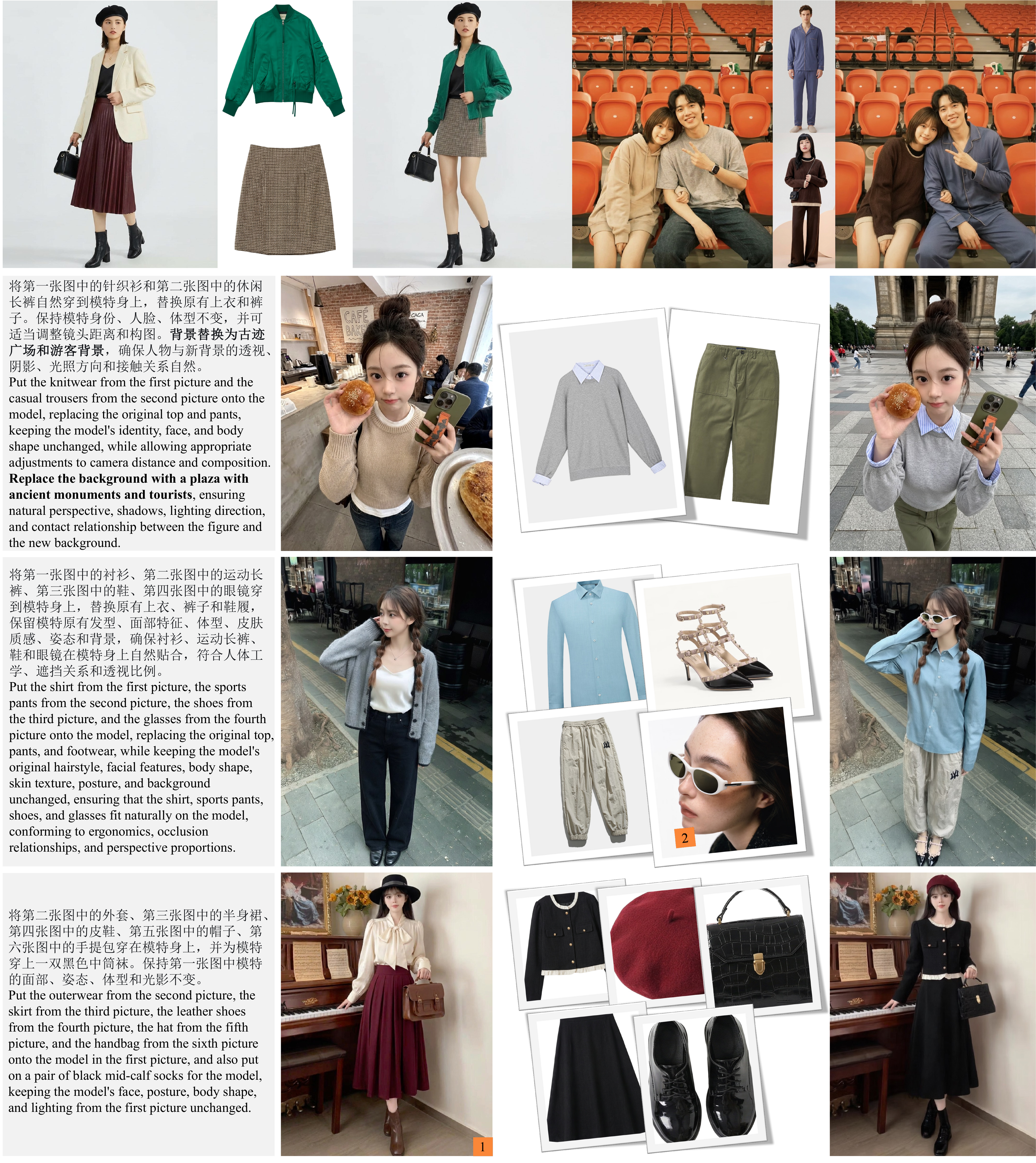}
    \caption{\textbf{Qualitative gallery of Oxygen-TryOn on multi-item try-on.} From multiple references provided at once, Oxygen-TryOn performs free multi-item composition with plausible layering and occlusion, dressing each subject in the intended combination while keeping every item's appearance and the subject's identity consistent.}
\label{fig:gallery_multi}
\end{figure}

\begin{figure}[H]
    \centering
    \includegraphics[width=\linewidth]{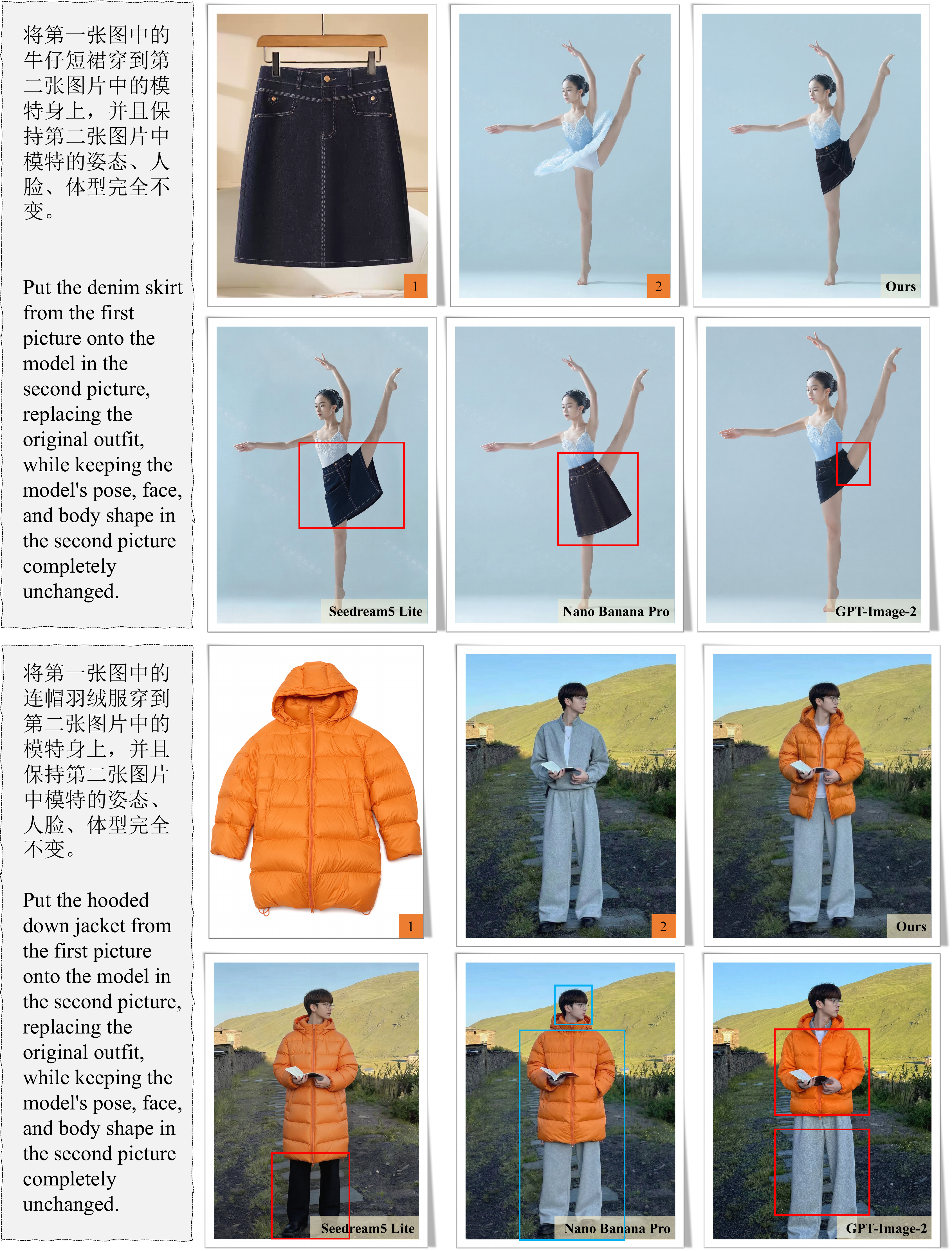}
    \caption{\textbf{Qualitative comparison with state-of-the-art proprietary systems on single-item try-on.}  \textcolor{red}{Red} boxes highlight local defects of the competing methods, and \textcolor{blue}{blue} boxes highlight implausible proportions (\eg, unnatural garment-to-body scale). Oxygen-TryOn better preserves item detail and yields well-proportioned results.}
\label{fig:cmp_single}
\end{figure}

Beyond the aggregate scores of Section~\ref{sec:quant}, we conduct a qualitative study that inspects, at the pixel level, where try-on quality holds up or breaks down. It is organized in two parts: open-ended capability demonstrations that probe the breadth of scenarios Oxygen-TryOn supports (Figures~\ref{fig:gallery_single} and~\ref{fig:gallery_multi}), and matched head-to-head comparisons against the strongest proprietary systems---Seedream5 Lite, Nano Banana Pro, and GPT-Image-2---under identical references and instructions (Figures~\ref{fig:cmp_single} and~\ref{fig:cmp_multi}). We read every result along three axes that jointly decide whether a try-on is deployable: (i)~\textbf{consistency}, faithful preservation of the person's identity, body shape, pose, and background together with each reference item's texture, pattern, structure, and logos; (ii)~\textbf{realism}, physically plausible drape, shadow, occlusion, and layering free of synthetic artifacts; and (iii)~\textbf{flexibility}, correct placement of every requested item and faithful execution of dense, multi-reference, and edit-augmented instructions.

\paragraph{Capability showcase.}
Figures~\ref{fig:gallery_single} and~\ref{fig:gallery_multi} map out the range of scenarios Oxygen-TryOn handles from a single reference and from several references at once. For \emph{single-reference} try-on (Figure~\ref{fig:gallery_single}), the gallery spans garments, footwear, and accessories under diverse poses, viewpoints, and capture conditions, driven equally by clean product shots and in-the-wild worn-on photos. A light-purple blouse with a pussy-bow neckline and a pale-blue lace-trimmed dress are transferred onto real subjects in outdoor scenes with fine fabric detail intact and each subject's face, posture, and surroundings preserved; footwear and sportswear behave just as naturally, with beige slouch boots completing a full-body street look and a metallic-silver Adidas track set draping over a seated subject; and a delicate silver pendant is added consistently across tightly cropped portraits with different necklines and head poses. Crucially, the model generalizes far beyond photographs of real people: it dresses a hand-drawn cartoon character in a zip-up hoodie while respecting the original line art, composites a tan utility jacket and jeans onto the classical oil portrait \emph{The Laughing Cavalier} and a yellow--green windbreaker with a leopard-print skirt onto a second painted subject without overwriting the brushwork or lighting, and transfers a grey Ami cardigan and a leopard-print Adidas track jacket onto cinematic film stills while matching their grain and color grading. For \emph{multi-reference} try-on (Figure~\ref{fig:gallery_multi}), the cases escalate from simple two-item replacement to dense, instruction-driven outfit assembly. A green satin bomber and a brown checkered skirt jointly replace a subject's top and bottom with plausible waist-level layering, and in a stadium scene two people are simultaneously re-dressed in a blue pajama set and a brown loungewear set while their relative pose and venue remain unchanged. Harder cases fuse wearing with editing in a single pass: guided by an explicit instruction, the model puts a grey knit top and olive trousers on the subject while replacing the original cafe interior with a monument plaza whose perspective, shadows, and ground contact stay coherent. At the highest cardinality, it composes four heterogeneous references---a blue shirt, grey joggers, strappy heels, and white-framed sunglasses---onto an outdoor portrait with correct per-region placement, and assembles a full six-element look---a black cardigan, maroon beret, crocodile-texture handbag, black skirt, lace-up shoes, and black mid-calf socks---beside a piano while resolving hat--hair occlusion, hand--bag interaction, and shoe--floor contact. Taken together, these results indicate that Oxygen-TryOn learns generalizable item--subject wearing relations across categories, visual domains, and reference cardinalities rather than a narrow studio-pose prior.

\paragraph{Comparison with proprietary systems.}
Figures~\ref{fig:cmp_single} and~\ref{fig:cmp_multi} place Oxygen-TryOn beside the strongest proprietary editing systems under identical references and instructions, progressing from single-item transfer to increasingly dense multi-item composition; following the caption convention, \textcolor{red}{red} boxes flag local appearance defects and \textcolor{blue}{blue} boxes flag implausible proportions.

\emph{Single-item transfer} (Figure~\ref{fig:cmp_single}). In the first case a dark denim skirt is placed on a ballet dancer holding a high leg extension: our model drapes the skirt over the raised leg with believable folds and hem curvature, whereas Seedream5 Lite renders it stiff and misaligned with the pose, Nano Banana Pro lays down a flat overlay that ignores the leg geometry, and GPT-Image-2 shortens it into a mini-skirt that no longer matches the reference length. In the second case a long orange hooded down jacket is transferred onto a man on a country path; Oxygen-TryOn fits the coat at its natural length while leaving the grey trousers and background untouched, but the baselines recolor the trousers, distort overall body proportions, or truncate the coat into a waist-length puffer---each breaking either item fidelity or the instruction to change only the specified garment.

\emph{Multi-item composition} (Figure~\ref{fig:cmp_multi}). In the first case, a dark pleated skirt with a brown belt and a white off-shoulder ruffled blouse are transferred onto a woman walking across a city crosswalk: Oxygen-TryOn swaps in all three items while preserving her face, hairstyle, gait, the iced drink in her hand, her green socks, shoulder bag, and street background, whereas the baselines each break a local detail---Seedream5 Lite distorts the off-shoulder blouse, while Nano Banana Pro and GPT-Image-2 mishandle the belt and waistline where the skirt meets the original shoulder bag. In the second case, a yellow knit headband and a maroon fleece  zip-up jacket are added to a model in a demanding forward-leaning pose as she holds a snack bag beside a white tote: Oxygen-TryOn keeps her pose, facial identity, body shape, the bag, and the blue-sky background intact, whereas Seedream5 Lite drifts on both the face and the jacket and distorts the overall proportions,  and Nano Banana Pro and GPT-Image-2 still shift the model's facial identity around the newly added headband.
 Mirroring the quantitative gaps in Section~\ref{sec:quant}, these   comparisons show that proprietary systems
 can look convincing on isolated edits, yet Oxygen-TryOn more reliably satisfies the joint constraints---identity preservation, per-item fidelity, and physically plausible multi-item placement---that real try-on deployments demand.

%% file: sections/radar.tex
\definecolor{oxyred}{RGB}{208,32,44}
\definecolor{fluxblue}{RGB}{31,95,200}
\definecolor{gptgreen}{RGB}{34,139,34}
\definecolor{nbporange}{RGB}{230,145,20}
\definecolor{seedpurple}{RGB}{124,58,160}

\begin{figure}[t]
\centering
\begin{subfigure}{0.46\linewidth}
\centering
\resizebox{\linewidth}{!}{%
\begin{tikzpicture}[
    font=\scriptsize,
    gridline/.style={draw=black!16, line width=0.4pt},
    axisline/.style={draw=black!35, line width=0.5pt},
    ax/.style={font=\scriptsize, align=center, text=black!72},
    tick/.style={font=\tiny, text=black!50, fill=white, fill opacity=0.85, text opacity=1, inner sep=0.6pt},
    oxy/.style={draw=oxyred, line width=1.3pt, fill=oxyred, fill opacity=0.10},
    flux/.style={draw=fluxblue, line width=0.9pt},
    gpt/.style={draw=gptgreen, line width=0.9pt},
    nbp/.style={draw=nbporange, line width=0.9pt},
    seed/.style={draw=seedpurple, line width=0.9pt}
]
% grid rings (scores 7,8,9,10), pentagon: Overall(top), Item, Phys, Backgr, Identity
\draw[gridline] (0,0.75)--(0.713,0.232)--(0.441,-0.607)--(-0.441,-0.607)--(-0.713,0.232)--cycle;
\draw[gridline] (0,1.5)--(1.427,0.464)--(0.882,-1.214)--(-0.882,-1.214)--(-1.427,0.464)--cycle;
\draw[gridline] (0,2.25)--(2.140,0.695)--(1.323,-1.820)--(-1.323,-1.820)--(-2.140,0.695)--cycle;
\draw[gridline] (0,3.0)--(2.853,0.927)--(1.764,-2.427)--(-1.764,-2.427)--(-2.853,0.927)--cycle;
\draw[draw=black!32, line width=0.6pt] (0,0) circle (3.0);
% axes
\draw[axisline] (0,0)--(0,3.0);
\draw[axisline] (0,0)--(2.853,0.927);
\draw[axisline] (0,0)--(1.764,-2.427);
\draw[axisline] (0,0)--(-1.764,-2.427);
\draw[axisline] (0,0)--(-2.853,0.927);
% --- single-item polygons (Overall, Item, Phys, Backgr, Identity) ---
\draw[flux] (0,1.32)--(0.392,0.127)--(0.732,-1.007)--(-1.103,-1.517)--(-1.783,0.579)--cycle;
\draw[gpt]  (0,1.755)--(1.791,0.582)--(0.975,-1.341)--(-1.085,-1.493)--(-2.496,0.811)--cycle;
\draw[nbp]  (0,1.53)--(1.891,0.614)--(0.940,-1.293)--(-0.829,-1.141)--(-2.247,0.730)--cycle;
\draw[seed] (0,2.078)--(1.862,0.605)--(1.094,-1.505)--(-1.407,-1.936)--(-2.618,0.851)--cycle;
\draw[oxy]  (0,2.52)--(2.211,0.718)--(1.451,-1.996)--(-1.627,-2.239)--(-2.725,0.885)--cycle;
\foreach \p in {(0,2.52),(2.211,0.718),(1.451,-1.996),(-1.627,-2.239),(-2.725,0.885)}{\fill[oxyred] \p circle (1.2pt);}
% radial tick labels (drawn last, white background above polygons)
\node[tick,anchor=east] at (-0.07,0.75){7};
\node[tick,anchor=east] at (-0.07,1.5){8};
\node[tick,anchor=east] at (-0.07,2.25){9};
\node[tick,anchor=east] at (-0.07,3.0){10};
% axis labels
\node[ax,anchor=south] at (0,3.2){Overall};
\node[ax,anchor=west]  at (2.95,0.96){Item\\Fidelity};
\node[ax,anchor=north] at (1.9,-2.6){Phys. \&\\Struc. Logic};
\node[ax,anchor=north] at (-1.9,-2.6){Backgr.\\Preserv.};
\node[ax,anchor=east]  at (-2.95,0.96){Identity\\Consist.};
\end{tikzpicture}%
}
\caption{Single-item try-on}
\label{fig:radar_single}
\end{subfigure}
\hfill
\begin{subfigure}{0.46\linewidth}
\centering
\resizebox{\linewidth}{!}{%
\begin{tikzpicture}[
    font=\scriptsize,
    gridline/.style={draw=black!16, line width=0.4pt},
    axisline/.style={draw=black!35, line width=0.5pt},
    ax/.style={font=\scriptsize, align=center, text=black!72},
    tick/.style={font=\tiny, text=black!50, fill=white, fill opacity=0.85, text opacity=1, inner sep=0.6pt},
    oxy/.style={draw=oxyred, line width=1.3pt, fill=oxyred, fill opacity=0.10},
    flux/.style={draw=fluxblue, line width=0.9pt},
    gpt/.style={draw=gptgreen, line width=0.9pt},
    nbp/.style={draw=nbporange, line width=0.9pt},
    seed/.style={draw=seedpurple, line width=0.9pt}
]
% grid rings (scores 4,6,8,10), axis scale 3--10
\draw[gridline] (0,0.4286)--(0.4076,0.1324)--(0.2520,-0.3467)--(-0.2520,-0.3467)--(-0.4076,0.1324)--cycle;
\draw[gridline] (0,1.2857)--(1.2227,0.3973)--(0.7560,-1.0401)--(-0.7560,-1.0401)--(-1.2227,0.3973)--cycle;
\draw[gridline] (0,2.1429)--(2.0379,0.6622)--(1.2600,-1.7336)--(-1.2600,-1.7336)--(-2.0379,0.6622)--cycle;
\draw[gridline] (0,3.0)--(2.853,0.927)--(1.764,-2.427)--(-1.764,-2.427)--(-2.853,0.927)--cycle;
\draw[draw=black!32, line width=0.6pt] (0,0) circle (3.0);
% axes
\draw[axisline] (0,0)--(0,3.0);
\draw[axisline] (0,0)--(2.853,0.927);
\draw[axisline] (0,0)--(1.764,-2.427);
\draw[axisline] (0,0)--(-1.764,-2.427);
\draw[axisline] (0,0)--(-2.853,0.927);
% --- multi-item (up to 4 conditions) polygons; axis scale 3--10, r=(s-3)*3/7 ---
\draw[flux] (0,0.681)--(0.049,0.016)--(0.262,-0.361)--(-1.023,-1.408)--(-1.928,0.626)--cycle;
\draw[gpt]  (0,2.100)--(2.172,0.706)--(1.187,-1.633)--(-1.308,-1.799)--(-2.458,0.799)--cycle;
\draw[nbp]  (0,2.010)--(2.242,0.728)--(1.222,-1.682)--(-1.154,-1.588)--(-2.291,0.744)--cycle;
\draw[seed] (0,2.224)--(2.238,0.727)--(1.283,-1.765)--(-1.436,-1.976)--(-2.356,0.765)--cycle;
\draw[oxy]  (0,2.319)--(2.050,0.666)--(1.404,-1.931)--(-1.512,-2.080)--(-2.474,0.804)--cycle;
\foreach \p in {(0,2.319),(2.050,0.666),(1.404,-1.931),(-1.512,-2.080),(-2.474,0.804)}{\fill[oxyred] \p circle (1.2pt);}
% radial tick labels (drawn last, white background above polygons)
\node[tick,anchor=east] at (-0.07,0.4286){4};
\node[tick,anchor=east] at (-0.07,1.2857){6};
\node[tick,anchor=east] at (-0.07,2.1429){8};
\node[tick,anchor=east] at (-0.07,3.0){10};
% axis labels
\node[ax,anchor=south] at (0,3.2){Overall};
\node[ax,anchor=west]  at (2.95,0.96){Item\\Fidelity};
\node[ax,anchor=north] at (1.9,-2.6){Phys. \&\\Struc. Logic};
\node[ax,anchor=north] at (-1.9,-2.6){Backgr.\\Preserv.};
\node[ax,anchor=east]  at (-2.95,0.96){Identity\\Consist.};
\end{tikzpicture}%
}
\caption{Multi-item try-on}
\label{fig:radar_multi}
\end{subfigure}

\vspace{1.5mm}
% shared legend
\begin{tikzpicture}[font=\scriptsize]
\draw[oxyred,line width=1.3pt] (0,0)--(0.55,0);
\fill[oxyred] (0.275,0) circle (1.2pt);
\node[anchor=west] at (0.62,0) {Oxygen-TryOn (Ours)};
\draw[fluxblue,line width=0.9pt] (4.0,0)--(4.55,0);
\node[anchor=west] at (4.62,0) {FLUX.2-dev};
\draw[gptgreen,line width=0.9pt] (6.3,0)--(6.85,0);
\node[anchor=west] at (6.92,0) {GPT-Image-2};
\draw[nbporange,line width=0.9pt] (9.3,0)--(9.85,0);
\node[anchor=west] at (9.92,0) {Nano Banana Pro};
\draw[seedpurple,line width=0.9pt] (12.6,0)--(13.15,0);
\node[anchor=west] at (13.22,0) {Seedream5 Lite};
\end{tikzpicture}
\caption{\textbf{Performance on single-item and multi-item scenarios on TStars-VTON dataset.} 
Each axis indicates a critical dimension for visual try-on. All scores are judged by powerful MLLMs and the \emph{Overall} score is aggregated by average. Oxygen-TryOn attains the most balanced, leading profile on \textbf{(a)} single-item try-on and remains competitive on \textbf{(b)} the demanding multi-item setting. See Tables~\ref{tab:single_results} and~\ref{tab:multi_results}.}
\label{fig:teaser_radar}
\end{figure}

%% file: sections/application.tex
\section{Applications and Capabilities}
\label{sec:applications}

Beyond standard single-item try-on, Oxygen-TryOn supports a range of practical capabilities that arise from its unified, understanding-driven design. We highlight three below.

\subsection{Free Multi-Item Composition}

Oxygen-TryOn accepts an arbitrary number of reference items and composes them into a single coherent outfit on the subject. This includes mixing categories, such as a top, bottom, coat, hat, shoes, and a bag in one pass, and resolving the layering and occlusion relationships among them. As shown in Figure~\ref{fig:gallery_multi}, the model dresses the subject with multiple items while preserving each item's appearance and the subject's identity, pose, and background. This free-combination ability is what enables full-outfit ``OOTD'' (outfit-of-the-day) style exploration rather than item-by-item replacement.

\subsection{Built-in General Editing}
\label{sec:editing}

Because Oxygen-TryOn is initialized from a general-purpose foundation model and trained with a mixture that includes editing-augmented try-on samples (Section~\ref{sec:sft}), it retains general instruction-based editing ability. Concretely, while generating the try-on result the model can additionally follow editing directives expressed in the instruction $T$, such as adjusting the subject's pose, without invoking a separate editing model or a second pass. This makes the system more flexible in practice, since users can refine the wearing result and the presentation jointly. We regard this as a useful built-in capability rather than the primary focus of the model, and we therefore demonstrate it succinctly in Figure~\ref{fig:gallery_multi}.

\subsection{Cross-Domain Generalization}

The understanding-driven design lets Oxygen-TryOn generalize beyond standard real-person photography. It can apply garments and accessories to non-standard subjects, such as stylized 3D avatars, illustrated characters, or statues, while respecting the original artistic style and geometry, as illustrated in Figure~\ref{fig:gallery_single}. This indicates that the model learns generalizable representations of items and wearing relations rather than overfitting to a narrow human-pose prior.

%% file: sections/related_work.tex
\section{Related Work}
\label{sec:related_work}

\subsection{Image-Based Virtual Try-On}
\label{sec:related_task_specific}

Image-based virtual try-on is classically posed as constrained person-image synthesis: a target person is recombined with a single reference garment under strong assumptions on body pose, garment category, and input layout. GAN-era systems such as VITON and CP-VTON established the dominant recipe of warping the garment to the body and compositing it while preserving the person's identity~\cite{han2018viton,wang2018toward}. Diffusion-based methods then sharply improved realism and detail preservation: IDM-VTON, OOTDiffusion, CatVTON, FitDiT, Leffa, and PromptDresser refine the conditioning, warping, masking, or attention used to fuse garment and person, and report strong numbers on VITON-HD and DressCode~\cite{choi2024idmvton,xu2025ootdiffusion,chong2024catvtonconcatenationneedvirtual,jiang2024fitdit,zhou2024learning,kim2024promptdresser}. Yet the bulk of this line stays centered on garment replacement under standardized inputs---typically a clean flat-lay garment, a studio person, and a garment-agnostic mask---which leaves little room for accessories, shoes, bags, or free multi-item composition.

A more recent line begins to loosen these constraints. Wear-Any-Way and Tunnel Try-on pursue more controllable and video-oriented try-on~\cite{chen2024wear,xu2024tunnel}, while Any2AnyTryon and FastFit move toward more versatile and multi-reference garment try-on~\cite{guo2025any2anytryon,chong2025fastfitacceleratingmultireferencevirtual}. These efforts are the closest in spirit to ours, yet they remain largely garment-centric and do not jointly handle heterogeneous references (clean product shots versus in-the-wild worn-on photos), non-garment items, and arbitrary multi-item outfits within a single model. Oxygen-TryOn departs from the inpainting/warping view and casts try-on as an understanding-driven, multi-reference generation task over arbitrary fashion items: it drops the garment-agnostic mask and lets one generation pass reason jointly about item identity, body-region placement, layering, and optional edits.

\subsection{General-Purpose Generation and Editing Models}
\label{sec:related_general_models}

Try-on systems are increasingly built on top of large general-purpose generators rather than bespoke warping pipelines. Denoising diffusion, latent diffusion, and rectified-flow transformers provide scalable, high-fidelity backbones~\cite{ho2020denoising,rombach2022high,esser2024scaling, xue2025retouchgpt,wen2024retouchformer}, and multimodal instruction-following models---open-source Qwen-Image, FLUX.2, and HunyuanImage~\cite{wu2025qwen,blackforest2025flux2klein,cao2025hunyuanimage}, together with strong proprietary systems such as Nano Banana Pro, GPT-Image-2, and Seedream5 Lite~\cite{google2025nanobanana,gptimage2_model_card,bytedance2026seedream}---pair photorealistic synthesis with broad semantic understanding. Because they parse open-ended prompts and need no task-specific warping, they can be prompted directly for try-on-like editing and serve as strong initializations for task-specific fine-tuning.

Their generality, however, is not tuned to the fine-grained consistency that try-on demands: a usable result must hold the person's identity, pose, body shape, and background fixed while exactly reproducing every reference item's texture, print, structure, and logo, and must keep multiple items physically coherent under layering and occlusion. Deviations that are harmless in open-ended editing become outright failures in fashion, and---as our experiments confirm---even strong general models that are competitive on single-item try-on degrade once several references must be honored at once. Oxygen-TryOn therefore takes such a general multimodal model as its foundation but specializes it into a fashion-native system through a dedicated data engine, multi-reference conditioning, and try-on-specific SFT and RL.

\subsection{Toward Open and Reproducible Try-On}
\label{sec:related_closed_source}

In practice, the most capable try-on behavior today is delivered by the proprietary systems noted above: prompted directly, they already produce polished, instruction-faithful edits on unconstrained product images and in-the-wild references, which makes them indispensable comparators for any deployable try-on model. Their training data, architecture, optimization recipe, and evaluation protocol are nevertheless undisclosed, so the community cannot diagnose \emph{why} a system corrupts a logo, drops a garment, drifts on identity, or produces implausible layering---precisely the failure modes that matter most for fashion. Open-source efforts, meanwhile, are mostly the task-specific try-on models discussed above, which remain confined to narrow settings and still trail closed systems in realistic use.

Oxygen-TryOn is positioned to close this gap from the open side: it aims to match strong closed-source quality in realistic, any-item try-on while releasing the model weights together with the data-engine design, the CPT--SFT--RL recipe, and the inference and prompt-enhancer protocol, so that the entire pipeline can be reproduced, audited, and built upon.

%% file: sections/limitation.tex
\section{Limitations and Future Work}
\label{sec:limitation}

Although Oxygen-TryOn delivers high-fidelity, any-item try-on across a wide range of single- and multi-item scenarios, it still has limitations that we plan to address in future work.

\noindent\textbf{Scaling to more reference items.} Oxygen-TryOn is built on a foundation model pretrained for at most four reference images, so we focus its multi-item use and evaluation on scenarios with up to four references, where the model operates within its primary capability range. Composing five or more references under complex layering and occlusion is substantially harder: as the number of items grows, the model becomes more prone to item confusion, dropped items, and implausible layering. We plan to lift this limit by scaling and rebalancing the data engine toward dense, high-cardinality multi-item outfits, strengthening the multi-reference conditioning, and moving to a base model with larger capacity, so that a single pass can faithfully assemble a full outfit from many references.

\noindent\textbf{Distilled, accelerated variant.} Like most diffusion-based generators, Oxygen-TryOn relies on multiple denoising steps at inference, which limits throughput for interactive and large-scale deployment. To make the model more practical, we will release a distilled, accelerated variant that markedly reduces the number of sampling steps while preserving item and subject consistency, providing faster and cheaper try-on with minimal loss in quality.

%% file: sections/conclusion.tex
\section{Conclusion}

In this report, we presented \textbf{Oxygen-TryOn}, a unified foundation model for any-item virtual try-on. Rather than casting try-on as mask-based inpainting, we reformulated it as a multi-reference, understanding-driven generation task: given heterogeneous reference items---clean product shots or in-the-wild worn-on photos---together with a single target subject image, Oxygen-TryOn synthesizes photorealistic results that faithfully preserve both item appearance and subject identity across single-item and multi-item, full-body and half-body scenarios. Built on the JoyAI-Image-Edit architecture and initialized from its pretrained weights, the model couples an MLLM with an MMDiT so that semantic understanding of the references and the wearing intent directly conditions high-fidelity synthesis. We unlocked this capability through a dedicated data engine that collects, manufactures, annotates, and filters high-quality try-on data, and a three-stage recipe combining continued pre-training, large-scale supervised fine-tuning, and reinforcement learning under a hybrid reward that fuses an in-house try-on reward model with a rubric-guided Gemini~3.1~Pro judge. Beyond dressing the target subject, Oxygen-TryOn retains general instruction-based editing, allowing the wearing result to be refined---for example, via a pose change---within the same generation pass.

Across public benchmarks and our Oxygen-TryOn Bench, Oxygen-TryOn attains state-of-the-art consistency and realism on single-item try-on, surpassing strong proprietary systems such as Nano Banana Pro, GPT-Image-2, and Seedream5 Lite, as well as leading open-source models such as FLUX.2, while also leading multi-item composition. To the best of our knowledge, it is the first system to deliver any-item, multi-reference try-on at this fidelity. We detail the data engine and CPT--SFT--RL recipe behind the model to make our design choices transparent. We hope that presenting Oxygen-TryOn, together with a transparent account of how it is built, helps move virtual try-on from constrained, single-category prototypes toward truly universal and broadly accessible systems.

%% file: sections/acknowledge.tex
% \clearpage
\renewcommand{\thefootnote}{\fnsymbol{footnote}}

\section{Authors}
\label{sec:contributions}

\textbf{Core Contributors}\\[0.5em]
Yong Liu\footnotemark[1], Xiaolong Fu\footnotemark[1], Zihang Xu, Wen Xue, Xueheng Li, Lin Song, Yuan Zhang, Chuyang Zhao, Haoyang Huang, Nan Duan, Yipeng~Sun\footnotemark[2], Yan Li, Simiu Gu.

\footnotetext[1]{Equal contribution.}
\footnotetext[2]{Corresponding author: Yipeng Sun <sunyipeng.luka@jd.com>.}

% \begingroup
% \textbf{Contributors}\footnote{Contributors are listed in alphabetical order.}
\textbf{Contributors}%\footnotemark[3]\\[0.5em]
% \endgroup \\

%\footnotetext[3]{Contributors are listed in alphabetical order.}

Shu Duan, Ting Zhu, Yichun Liu, Peng Liu, Dajiang Wu, Qili Wang, Tao Yuan, Zhuofan Xia, Xinyu Gong, Dongsheng Jia, Ruofan Lv, Maoquan Zhang, Xin Han, Wei Tang, Jiachen Liu, Chengzhi Huang, Yang Pei, Shaolong Xing, Zhen Chen, Ke Zhang.

%% file: paper.bbl
\begin{thebibliography}{47}
\providecommand{\natexlab}[1]{#1}
\providecommand{\url}[1]{\texttt{#1}}
\expandafter\ifx\csname urlstyle\endcsname\relax
  \providecommand{\doi}[1]{doi: #1}\else
  \providecommand{\doi}{doi: \begingroup \urlstyle{rm}\Url}\fi

\bibitem[Bai et~al.(2025{\natexlab{a}})Bai, Cai, Chen, Huang, Li, Lin, Zhu, et~al.]{qwen3vl2025}
Shuai Bai, Yuxuan Cai, Xionghui Chen, Qidong Huang, Kaixin Li, Zicheng Lin, Keming Zhu, et~al.
\newblock Qwen3-vl technical report.
\newblock \emph{arXiv preprint arXiv:2511.21631}, 2025{\natexlab{a}}.
\newblock URL \url{https://arxiv.org/abs/2511.21631}.

\bibitem[Bai et~al.(2025{\natexlab{b}})Bai, Chen, Liu, Wang, Ge, Song, Dang, Wang, Wang, Tang, et~al.]{qwen2.5vl}
Shuai Bai, Keqin Chen, Xuejing Liu, Jialin Wang, Wenbin Ge, Sibo Song, Kai Dang, Peng Wang, Shijie Wang, Jun Tang, et~al.
\newblock Qwen2. 5-vl technical report.
\newblock \emph{arXiv:2502.13923}, 2025{\natexlab{b}}.

\bibitem[{Black Forest Labs}(2025)]{blackforest2025flux2klein}
{Black Forest Labs}.
\newblock Flux.2: Towards interactive visual intelligence, 2025.
\newblock URL \url{https://bfl.ai/blog/flux2}.

\bibitem[ByteDance(2026)]{bytedance2026seedream}
ByteDance.
\newblock Deeper thinking, more accurate generation: Introducing seedream 5.0 lite, 2026.
\newblock URL \url{https://seed.bytedance.com/en/blog/}.

\bibitem[Cao et~al.(2025)Cao, Chen, Chen, et~al.]{cao2025hunyuanimage}
Siyu Cao, Hangting Chen, Peng Chen, et~al.
\newblock Hunyuanimage 3.0 technical report.
\newblock \emph{arXiv preprint arXiv:2509.23951}, 2025.

\bibitem[Chen et~al.(2024)Chen, Chen, Zhai, Ju, Hong, Lan, and Xiao]{chen2024wear}
Mengting Chen, Xi~Chen, Zhonghua Zhai, Chen Ju, Xuewen Hong, Jinsong Lan, and Shuai Xiao.
\newblock Wear-any-way: Manipulable virtual try-on via sparse correspondence alignment.
\newblock In \emph{European Conference on Computer Vision}, pages 124--142. Springer, 2024.

\bibitem[Chen et~al.(2026)Chen, Chen, Du, Gao, Hu, Lan, Lin, Shen, Wang, Wang, Wu, Xu, Xu, Yan, Zhang, Zheng, Zhou, Zhu, and Zheng]{chen2026tstars}
Mengting Chen, Zhengrui Chen, Yongchao Du, Zuan Gao, Taihang Hu, Jinsong Lan, Chao Lin, Yefeng Shen, Xingjian Wang, Zhao Wang, Zhengtao Wu, Xiaoli Xu, Zhengze Xu, Hao Yan, Mingzhou Zhang, Jun Zheng, Qinye Zhou, Xiaoyong Zhu, and Bo~Zheng.
\newblock Tstars-tryon 1.0: Robust and realistic virtual try-on for diverse fashion items, 2026.

\bibitem[Choi et~al.(2021)Choi, Park, Lee, and Choo]{choi2021vitonhd}
Seunghwan Choi, Sunghyun Park, Minsoo Lee, and Jaegul Choo.
\newblock Viton-hd: High-resolution virtual try-on via misalignment-aware normalization.
\newblock In \emph{Proceedings of the IEEE/CVF Conference on Computer Vision and Pattern Recognition}, pages 14131--14140, 2021.

\bibitem[Choi et~al.(2024)Choi, Kwak, Lee, Choi, and Shin]{choi2024idmvton}
Yisol Choi, Sangkyung Kwak, Kyungmin Lee, Hyungwon Choi, and Jinwoo Shin.
\newblock Improving diffusion models for authentic virtual try-on in the wild.
\newblock In \emph{European Conference on Computer Vision (ECCV)}, 2024.

\bibitem[Chong et~al.(2024)Chong, Dong, Li, Zhang, Zhang, Zhang, Zhao, and Liang]{chong2024catvtonconcatenationneedvirtual}
Zheng Chong, Xiao Dong, Haoxiang Li, Shiyue Zhang, Wenqing Zhang, Xujie Zhang, Hanqing Zhao, and Xiaodan Liang.
\newblock Catvton: Concatenation is all you need for virtual try-on with diffusion models, 2024.
\newblock URL \url{https://arxiv.org/abs/2407.15886}.

\bibitem[Chong et~al.(2025)Chong, Lei, Zhang, He, Wang, Zhang, Dong, Wu, Jiang, and Liang]{chong2025fastfitacceleratingmultireferencevirtual}
Zheng Chong, Yanwei Lei, Shiyue Zhang, Zhuandi He, Zhen Wang, Xujie Zhang, Xiao Dong, Yiling Wu, Dongmei Jiang, and Xiaodan Liang.
\newblock Fastfit: Accelerating multi-reference virtual try-on via cacheable diffusion models, 2025.
\newblock URL \url{https://arxiv.org/abs/2508.20586}.

\bibitem[Cui et~al.(2025)Cui, Sun, Lin, Gao, Zhang, Liu, Wang, Zhang, Zhou, Liu, Zhang, Lv, Huang, Zhang, Zhang, Zhang, Liu, Yu, and Ma]{cui2025paddleocr30technicalreport}
Cheng Cui, Ting Sun, Manhui Lin, Tingquan Gao, Yubo Zhang, Jiaxuan Liu, Xueqing Wang, Zelun Zhang, Changda Zhou, Hongen Liu, Yue Zhang, Wenyu Lv, Kui Huang, Yichao Zhang, Jing Zhang, Jun Zhang, Yi~Liu, Dianhai Yu, and Yanjun Ma.
\newblock Paddleocr 3.0 technical report, 2025.
\newblock URL \url{https://arxiv.org/abs/2507.05595}.

\bibitem[Cui et~al.(2026)Cui, Sun, Liang, Gao, Zhang, Liu, Wang, Zhou, Liu, Lin, et~al.]{cui2026paddleocr}
Cheng Cui, Ting Sun, Suyin Liang, Tingquan Gao, Zelun Zhang, Jiaxuan Liu, Xueqing Wang, Changda Zhou, Hongen Liu, Manhui Lin, et~al.
\newblock Paddleocr-vl-1.5: Towards a multi-task 0.9 b vlm for robust in-the-wild document parsing.
\newblock \emph{arXiv preprint arXiv:2601.21957}, 2026.

\bibitem[discus0434(2024)]{discus2024aestheticpredictorv25}
discus0434.
\newblock Aesthetic predictor v2.5.
\newblock \url{https://github.com/discus0434/aesthetic-predictor-v2-5}, 2024.
\newblock SigLIP-based aesthetic score predictor. Accessed: 2026-06-30.

\bibitem[Esser et~al.(2024)Esser, Kulal, Blattmann, Entezari, M{\"u}ller, Saini, Levi, Lorenz, Sauer, Boesel, et~al.]{esser2024scaling}
Patrick Esser, Sumith Kulal, Andreas Blattmann, Rahim Entezari, Jonas M{\"u}ller, Harry Saini, Yam Levi, Dominik Lorenz, Axel Sauer, Frederic Boesel, et~al.
\newblock Scaling rectified flow transformers for high-resolution image synthesis.
\newblock In \emph{ICML}, 2024.

\bibitem[Fang et~al.(2020)Fang, Zhu, Zeng, Ma, and Wang]{fang2020perceptual}
Yuming Fang, Hanwei Zhu, Yan Zeng, Kede Ma, and Zhou Wang.
\newblock Perceptual quality assessment of smartphone photography.
\newblock In \emph{Proceedings of the IEEE/CVF conference on computer vision and pattern recognition}, pages 3677--3686, 2020.

\bibitem[Fu et~al.(2025)Fu, Ma, Guo, Zhou, Wang, Dong, Zhou, Liu, Fu, Sin, et~al.]{fu2025dynamic}
Xiaolong Fu, Lichen Ma, Zipeng Guo, Gaojing Zhou, Chongxiao Wang, ShiPing Dong, Shizhe Zhou, Ximan Liu, Jingling Fu, Tan~Lit Sin, et~al.
\newblock Dynamic-treerpo: Breaking the independent trajectory bottleneck with structured sampling.
\newblock \emph{arXiv preprint arXiv:2509.23352}, 2025.

\bibitem[{Google}(2025)]{google2025nanobanana}
{Google}.
\newblock Introducing nano banana pro, 2025.
\newblock URL \url{https://blog.google/innovation-and-ai/products/nano-banana-pro/}.
\newblock Accessed: 2026-05-30.

\bibitem[{Google}(2026)]{gemini31pro}
{Google}.
\newblock Gemini 3.1 pro: Announcing our latest gemini ai model.
\newblock \url{https://blog.google/innovation-and-ai/models-and-research/gemini-models/gemini-3-1-pro/}, February 2026.
\newblock Google Blog.

\bibitem[Guo et~al.(2025)Guo, Zeng, Song, Zhang, Liu, and Zhang]{guo2025any2anytryon}
Hailong Guo, Bohan Zeng, Yiren Song, Wentao Zhang, Jiaming Liu, and Chuang Zhang.
\newblock Any2anytryon: Leveraging adaptive position embeddings for versatile virtual clothing tasks.
\newblock In \emph{Proceedings of the IEEE/CVF International Conference on Computer Vision}, pages 19085--19096, 2025.

\bibitem[Han et~al.(2018)Han, Wu, Wu, Yu, and Davis]{han2018viton}
Xintong Han, Zuxuan Wu, Zhe Wu, Ruichi Yu, and Larry~S Davis.
\newblock Viton: An image-based virtual try-on network.
\newblock In \emph{Proceedings of the IEEE conference on computer vision and pattern recognition}, pages 7543--7552, 2018.

\bibitem[Ho et~al.(2020)Ho, Jain, and Abbeel]{ho2020denoising}
Jonathan Ho, Ajay Jain, and Pieter Abbeel.
\newblock Denoising diffusion probabilistic models.
\newblock \emph{Advances in neural information processing systems}, 33:\penalty0 6840--6851, 2020.

\bibitem[Hosu et~al.(2020)Hosu, Lin, Sziranyi, and Saupe]{hosu2020koniq}
Vlad Hosu, Hanhe Lin, Tamas Sziranyi, and Dietmar Saupe.
\newblock Koniq-10k: An ecologically valid database for deep learning of blind image quality assessment.
\newblock \emph{IEEE Transactions on Image Processing}, 29:\penalty0 4041--4056, 2020.

\bibitem[Jiang et~al.(2024)Jiang, Hu, Luo, He, Xu, Peng, Zhang, Wang, Wu, and Fu]{jiang2024fitdit}
Boyuan Jiang, Xiaobin Hu, Donghao Luo, Qingdong He, Chengming Xu, Jinlong Peng, Jiangning Zhang, Chengjie Wang, Yunsheng Wu, and Yanwei Fu.
\newblock Fitdit: Advancing the authentic garment details for high-fidelity virtual try-on.
\newblock \emph{arXiv preprint arXiv:2411.10499}, 2024.

\bibitem[Jin et~al.(2024)Jin, Qiao, Lu, Wang, Huang, Gao, Liu, and Li]{jin2024apddv2}
Xin Jin, Qianqian Qiao, Yi~Lu, Huaye Wang, Heng Huang, Shan Gao, Jianfei Liu, and Rui Li.
\newblock Apddv2: Aesthetics of paintings and drawings dataset with artist labeled scores and comments.
\newblock \emph{Advances in Neural Information Processing Systems}, 37:\penalty0 103064--103075, 2024.

\bibitem[Kim et~al.(2024)Kim, Jin, Park, and Choo]{kim2024promptdresser}
Jeongho Kim, Hoiyeong Jin, Sunghyun Park, and Jaegul Choo.
\newblock Promptdresser: Improving the quality and controllability of virtual try-on via generative textual prompt and prompt-aware mask.
\newblock \emph{arXiv preprint arXiv:2412.16978}, 2024.

\bibitem[Labs(2025)]{flux-2-2025}
Black~Forest Labs.
\newblock {FLUX.2: State-of-the-Art Visual Intelligence}.
\newblock \url{https://bfl.ai/blog/flux-2}, 2025.

\bibitem[Liu et~al.(2025)Liu, Liu, Liang, Li, Liu, Wang, Wan, Zhang, and Ouyang]{liu2025flow}
Jie Liu, Gongye Liu, Jiajun Liang, Yangguang Li, Jiaheng Liu, Xintao Wang, Pengfei Wan, Di~Zhang, and Wanli Ouyang.
\newblock Flow-grpo: Training flow matching models via online rl.
\newblock \emph{arXiv preprint arXiv:2505.05470}, 2025.

\bibitem[Ma et~al.(2025)Ma, Wu, Sun, and Li]{ma2025hpsv3}
Yuhang Ma, Xiaoshi Wu, Keqiang Sun, and Hongsheng Li.
\newblock Hpsv3: Towards wide-spectrum human preference score.
\newblock In \emph{Proceedings of the IEEE/CVF International Conference on Computer Vision}, pages 15086--15095, 2025.

\bibitem[Morelli et~al.(2022)Morelli, Fincato, Cornia, Landi, Cesari, and Cucchiara]{morelli2022dresscode}
Davide Morelli, Matteo Fincato, Marcella Cornia, Federico Landi, Fabio Cesari, and Rita Cucchiara.
\newblock Dress code: High-resolution multi-category virtual try-on.
\newblock In \emph{Computer Vision -- ECCV 2022}, pages 345--362, 2022.

\bibitem[OpenAI(2025)]{gptimage15_model_card}
OpenAI.
\newblock Gpt-image-1.5 model card, 2025.
\newblock URL \url{https://platform.openai.com/docs/models/gpt-image-1-5}.

\bibitem[OpenAI(2026)]{gptimage2_model_card}
OpenAI.
\newblock Gpt-image-2 model card, 2026.
\newblock URL \url{https://platform.openai.com/docs/models/gpt-image-2}.

\bibitem[Rombach et~al.(2022)Rombach, Blattmann, Lorenz, Esser, and Ommer]{rombach2022high}
Robin Rombach, Andreas Blattmann, Dominik Lorenz, Patrick Esser, and Bj{\"o}rn Ommer.
\newblock High-resolution image synthesis with latent diffusion models.
\newblock In \emph{Proceedings of the IEEE/CVF conference on computer vision and pattern recognition}, pages 10684--10695, 2022.

\bibitem[Shao et~al.(2024)Shao, Wang, Zhu, Xu, Song, Bi, et~al.]{shao2024deepseekmath}
Zhihong Shao, Peiyi Wang, Qihao Zhu, Runxin Xu, Junxiao Song, Xiao Bi, et~al.
\newblock Deepseekmath: Pushing the limits of mathematical reasoning in open language models.
\newblock \emph{arXiv preprint arXiv:2402.03300}, 2024.

\bibitem[Song et~al.(2026)Song, Li, Ma, Tang, Wang, Zhang, Yang, Xiao, Liu, Zhang, et~al.]{song2026joyai}
Lin Song, Wenbo Li, Guoqing Ma, Wei Tang, Bo~Wang, Yuan Zhang, Yijun Yang, Yicheng Xiao, Jianhui Liu, Yanbing Zhang, et~al.
\newblock Joyai-image: Awaking spatial intelligence in unified multimodal understanding and generation.
\newblock \emph{arXiv preprint arXiv:2605.04128}, 2026.

\bibitem[Team et~al.(2026)Team, Qiao, Hui, Li, Wang, Song, Zhang, Li, Xiang, Wang, et~al.]{team2026firered}
Super~Intelligence Team, Changhao Qiao, Chao Hui, Chen Li, Cunzheng Wang, Dejia Song, Jiale Zhang, Jing Li, Qiang Xiang, Runqi Wang, et~al.
\newblock Firered-image-edit-1.0 techinical report.
\newblock \emph{arXiv preprint arXiv:2602.13344}, 2026.

\bibitem[Wan et~al.(2025)Wan, Wang, Ai, Wen, Mao, Xie, Chen, Yu, Zhao, Yang, et~al.]{wan2025wan}
Team Wan, Ang Wang, Baole Ai, Bin Wen, Chaojie Mao, Chen-Wei Xie, Di~Chen, Feiwu Yu, Haiming Zhao, Jianxiao Yang, et~al.
\newblock Wan: Open and advanced large-scale video generative models.
\newblock \emph{arXiv preprint arXiv:2503.20314}, 2025.

\bibitem[Wang et~al.(2018)Wang, Zheng, Liang, Chen, Lin, and Yang]{wang2018toward}
Bochao Wang, Huabin Zheng, Xiaodan Liang, Yimin Chen, Liang Lin, and Meng Yang.
\newblock Toward characteristic-preserving image-based virtual try-on network.
\newblock In \emph{Proceedings of the European conference on computer vision (ECCV)}, pages 589--604, 2018.

\bibitem[Wen et~al.(2024)Wen, Xie, Jiang, Chen, Wu, Liu, and Wong]{wen2024retouchformer}
Xue Wen, Lianxin Xie, Le~Jiang, Tianyi Chen, Si~Wu, Cheng Liu, and Hau-San Wong.
\newblock Retouchformer: semi-supervised high-quality face retouching transformer with prior-based selective self-attention.
\newblock In \emph{Proceedings of the AAAI Conference on Artificial Intelligence}, volume~38, pages 5903--5911, 2024.

\bibitem[Wu et~al.(2025{\natexlab{a}})Wu, Li, Zhou, Lin, Gao, Yan, ming Yin, Bai, Xu, Chen, Chen, Tang, Zhang, Wang, Yang, Yu, Cheng, Liu, Li, Zhang, Meng, Wei, Ni, Chen, Cao, Peng, Qu, Wu, Wang, Yu, Wen, Feng, Xu, Wang, Zhang, Zhu, Wu, Cai, and Liu]{qwenimage}
Chenfei Wu, Jiahao Li, Jingren Zhou, Junyang Lin, Kaiyuan Gao, Kun Yan, Sheng ming Yin, Shuai Bai, Xiao Xu, Yilei Chen, Yuxiang Chen, Zecheng Tang, Zekai Zhang, Zhengyi Wang, An~Yang, Bowen Yu, Chen Cheng, Dayiheng Liu, Deqing Li, Hang Zhang, Hao Meng, Hu~Wei, Jingyuan Ni, Kai Chen, Kuan Cao, Liang Peng, Lin Qu, Minggang Wu, Peng Wang, Shuting Yu, Tingkun Wen, Wensen Feng, Xiaoxiao Xu, Yi~Wang, Yichang Zhang, Yongqiang Zhu, Yujia Wu, Yuxuan Cai, and Zenan Liu.
\newblock Qwen-image technical report.
\newblock \emph{arXiv preprint arXiv:2508.02324}, 2025{\natexlab{a}}.

\bibitem[Wu et~al.(2025{\natexlab{b}})Wu, Li, Zhou, Lin, et~al.]{wu2025qwen}
Chenfei Wu, Jiahao Li, Jingren Zhou, Junyang Lin, et~al.
\newblock Qwen-image technical report.
\newblock \emph{arXiv preprint arXiv:2508.02324}, 2025{\natexlab{b}}.

\bibitem[Wu et~al.(2025{\natexlab{c}})Wu, Jiang, Ku, Nie, Liu, and Chen]{wu2025editreward}
Keming Wu, Sicong Jiang, Max Ku, Ping Nie, Minghao Liu, and Wenhu Chen.
\newblock Editreward: A human-aligned reward model for instruction-guided image editing.
\newblock \emph{arXiv preprint arXiv:2509.26346}, 2025{\natexlab{c}}.

\bibitem[Xu et~al.(2025)Xu, Gu, Chen, and Chen]{xu2025ootdiffusion}
Yuhao Xu, Tao Gu, Weifeng Chen, and Arlene Chen.
\newblock Ootdiffusion: Outfitting fusion based latent diffusion for controllable virtual try-on.
\newblock In \emph{Proceedings of the AAAI Conference on Artificial Intelligence}, volume~39, pages 8996--9004, 2025.

\bibitem[Xu et~al.(2024)Xu, Chen, Wang, Xing, Zhai, Sang, Lan, Xiao, and Gao]{xu2024tunnel}
Zhengze Xu, Mengting Chen, Zhao Wang, Linyu Xing, Zhonghua Zhai, Nong Sang, Jinsong Lan, Shuai Xiao, and Changxin Gao.
\newblock Tunnel try-on: Excavating spatial-temporal tunnels for high-quality virtual try-on in videos.
\newblock In \emph{Proceedings of the 32nd ACM International Conference on Multimedia}, pages 3199--3208, 2024.

\bibitem[Xue et~al.(2025)Xue, Ding, Xu, Wu, Xu, and Wong]{xue2025retouchgpt}
Wen Xue, Chun Ding, Ruotao Xu, Si~Wu, Yong Xu, and Hau-San Wong.
\newblock Retouchgpt: Llm-based interactive high-fidelity face retouching via imperfection prompting.
\newblock In \emph{Proceedings of the AAAI Conference on Artificial Intelligence}, volume~39, pages 9059--9067, 2025.

\bibitem[Zheng et~al.(2025)Zheng, Chen, Ye, Wang, Zhang, Jiang, Su, Ermon, Zhu, and Liu]{zheng2025diffusionnft}
Kaiwen Zheng, Huayu Chen, Haotian Ye, Haoxiang Wang, Qinsheng Zhang, Kai Jiang, Hang Su, Stefano Ermon, Jun Zhu, and Ming-Yu Liu.
\newblock Diffusionnft: Online diffusion reinforcement with forward process.
\newblock \emph{arXiv preprint arXiv:2509.16117}, 2025.

\bibitem[Zhou et~al.(2024)Zhou, Liu, Han, Liu, Ng, Xie, Cong, Li, Xu, P{\'e}rez-R{\'u}a, Patel, Xiang, Shi, and He]{zhou2024learning}
Zijian Zhou, Shikun Liu, Xiao Han, Haozhe Liu, Kam~Woh Ng, Tian Xie, Yuren Cong, Hang Li, Mengmeng Xu, Juan-Manuel P{\'e}rez-R{\'u}a, Aditya Patel, Tao Xiang, Miaojing Shi, and Sen He.
\newblock Learning flow fields in attention for controllable person image generation.
\newblock \emph{arXiv preprint arXiv:2412.08486}, 2024.

\end{thebibliography}
